\documentclass{article}
\usepackage[utf8]{inputenc}
\usepackage{amsmath}
\usepackage{authblk}
\usepackage{comment}
\usepackage[numbers]{natbib}
\usepackage{graphics}
\usepackage{graphicx}
\usepackage{lipsum}

\title{A Two Parameters Equation \\ for Word Rank-Frequency Relation}
\author{Chenchen Ding }
\affil{
National Institute of Information and Communications Technology \\
3-5 Hikaridai, Seika-cho, Soraku-gun, Kyoto, 619-0289, Japan \\
{\tt chenchen.ding@nict.go.jp}}
\date{}

\begin{document}

\maketitle

\begin{abstract}
Let $f (\cdot)$ be the absolute frequency of words and $r$ be the rank of words in decreasing order of frequency, then the following function can fit the rank-frequency relation
\[
    f (r;s,t) = 
    \left(\frac{r_{\tt max}}{r}\right)^{1-s}
    \left(\frac{r_{\tt max}+t \cdot r_{\tt exp}}{r+t \cdot r_{\tt exp}}\right)^{1+(1+t)s}
\]
where $r_{\tt max}$ and $r_{\tt exp}$ are the maximum and the expectation of the rank, respectively; $s>0$ and $t>0$ are parameters estimated from data. On well-behaved data, there should be $s<1$ and $s \cdot t < 1$.   

\end{abstract}

\section{Introduction}

Zipf's law \cite{zipf1935psycho, zipf1949human} is an empirical law to formulate the rank-frequency relation in physical and social phenomena. Linguistically, Zipf's law can be observed on the distribution of words in corpora of natural languages, where the frequency ($f$) of words is inversely proportional to its rank ($r$) by frequency; that is, $f \propto r^{-1}$. Zipf's law is a special form of a general power law, i.e., $f \propto r^{-\alpha}$.

The Zipf's/power law is usually examined under a log-log plot of rank and frequency, where the data points lie on a straight line. The simple proportionality of the Zipf's/power law can be observed on randomly generated textual data~\cite{li1992random} and it only roughly depicts the rank-frequency relation in real textual data. A two-parameter generalization of the Zipf's/power law is the Zipf-Mandelbrot law, where $f \propto (r+\beta)^{-\alpha}$ \cite{mandelbrot1965information}.

\citet{li2010fitting} considered the reversed rank of $r_{\tt max} + 1 - r$, where $r_{\tt max}$ is the maximum of ranking index, and proposed $f \propto r^{-\alpha} (r_{\tt max} + 1 - r)^{\beta}$. This formulation is suitable for the distribution on letters, i.e., on symbols from a closed set.
\citet{ding2020three} proposed a three-parameter formulation of $f \propto r^{-\alpha}(r+\gamma)^{-\beta}$, which is derived based on observation and analysis of multilingual corpora. The Li's and Ding's formulations are in a form of the beta distribution of the first/second kind, respectively, regardless of the normalization constant. 

In this manuscript, we further introduce boundary and expectation values to reduce the parameter from four (three with the proportional coefficient) to two in Ding's formulation. The left two parameters depict the decreasing speed of the rank for frequent and obscure words. Experiments show that the two-parameter formulation can still fit the rank-frequency relation well.

\section{Derivation}

The beta distribution of the second kind, or the beta prime distribution has a probability density function of
\begin{equation}
    f (x; \alpha, \beta) =
    \frac{x^{\alpha-1}(1+x)^{-\alpha-\beta}}
    {B (\alpha, \beta)}
    \label{dist:betaprime}
\end{equation}
where $\alpha>0$, $\beta>0$, $B$ is the beta function.
It can be generalized to
\begin{equation}
    f(x; \alpha, \beta, p, q) =
    \frac{p\cdot (x/q)^{\alpha \cdot p-1}(1+(x/q)^p)^{-\alpha-\beta}}
    {q \cdot B (\alpha, \beta)}
\end{equation}
where $p>0$ and $q>0$ are two parameters for the shape and scale.

The Ding's formulation is in a form of the generalized beta prime distribution with $p=1$, i.e.,
\begin{equation}
    f(r; \alpha, \beta, 1, q) =
    \frac{r^{\alpha-1}(r+q)^{-\alpha-\beta}}
    {q^{\beta} \cdot B (\alpha, \beta)}
    \label{eq:gbp}
\end{equation}

The expectation of this distribution is $q\cdot\alpha/(\beta-1)$, when $\beta>1$. If the expectation $r_{\tt exp}$ is estimated by the rank-frequency pairs $(r_k, f_k)$ from the data, then we have
\begin{equation}
    r_{\tt exp} = \frac{\sum_k r_k \cdot f_k}{\sum_k f_k} =
    q \cdot \frac{\alpha}{\beta-1}
\end{equation}

By substituting $\alpha$, $\beta$, and $q$ by $s$, $1+s \cdot t$, and $t \cdot r_{\tt exp}$, respectively, and omitting the normalization term in \eqref{eq:gbp}, we have
\begin{equation}
    f (r;s,t) \propto 
    \left(\frac{1}{r}\right)^{1-s}
    \left(\frac{1}{r+t \cdot r_{\tt exp}}\right)^{1+(1+t)s}
    \label{propto:st}
\end{equation}

Considering the case of the least frequent words, whose frequency should be $1$ in most cases (i.e., the existence of singletons). Let their rank be $r_{\tt max}$ and the \eqref{propto:st} satisfies $f(r_{\tt max}; s, t) = 1$. We then have a formulation by an equation of
\begin{equation}
    f (r;s,t) = 
    \left(\frac{r_{\tt max}}{r}\right)^{1-s}
    \left(\frac{r_{\tt max}+t \cdot r_{\tt exp}}{r+t \cdot r_{\tt exp}}\right)^{1+(1+t)s}
    \label{eq:st}
\end{equation}
where $s >0$ and $t>0$ are parameters.
\section{Interpretation}

For many phenomena roughly following a power law, there is usually a well-defined mean but not such a variance \cite{newman2005power}, i.e., the second and higher order moments are undefined. This is also intuitive for the case of word rank-frequency. A well-defined mean means there is a relatively closed set of commonly used words; the nonexistence of variance means it is an open set for the vocabulary where any word may appear no matter how obscure it is. Therefore, if $f(r)$ is a function depicting the word rank-frequency relation, we hope that $\int f (r) dr$ and $\int rf(r) dr$ exist, but $\int r^k f (r) dr$ dose not exist when $k>2$. 

As to the original power law of $f \propto r^{-\alpha}$, it can be normalized to a distribution only when $\alpha >1$ and the distribution has moments up to the $k$-th when $\alpha > k+1$. To meet the above-mentioned properties, the $\alpha$ should be between $2$ and $3$. This is obvious too dramatic a dropping of the frequency against the rank. The problem can be attributed to the unified $\alpha$ on the entire vocabulary. From the experiments in \citet{ding2020three}, we can find the $\alpha$ is around $1$ for common words and around $2$ to $3$ for rare words on multilingual data. Therefore,  in reality the frequency concentrates more on common words, which is roughly Zipfian, and a fast dropping on rare words meets the properties required by a proper distribution.

The difference of Li's and Ding's formulations is also explainable. As the Li's formulation is in a form of the beta distribution (of the first kind), the \mbox{$k$-th} moments always exist once normalized. It is naturally suitable for rank-frequency relations on symbols from a closed set without unlimited obscure symbols. The Ding's formulation results in a huge slope for the rare symbols in such case, just to guarantee there are high order moments.

In the equation \eqref{eq:st}, the dropping slopes of common and rare words are parameterized by $1-s$ and $1+(1+t)s$. Notice here the $s$ provide a trade-off between the slopes of common and rare words. Considering a language tending to use multiple word expressions, rather than single words, for obscure concepts, it will have less rare words but a portion of common words will be more used for these obscure concepts. Under such case, the $s$ will turn larger to slow down the dropping of common words but to accelerate those rare words. The $t$ contributes to the dropping slope of the rare words and moreover, suggests where it begins to drop. The $t$ can be considered as a parameter depicting the portion of rare words from a corpus. The larger $t$, the smaller portion of rare words, i.e., from the rank of $t \cdot r_{\tt exp}$, the words can be considered as rare ones that accelerating to drop. Once $t > 1/s$, it turns to be able to define a variation for a normalized distribution on the vocabulary. This suggests the portion of rare words cannot go to infinitely small. A too large $t$ suggests that the symbol set may have less diversity than a vocabulary composed of ``words'' in general commonsense,\footnote{The vocabulary may be controlled, or the unit may be smaller than words, such as morphemes or characters.} for which the Li's formulation may be more proper to fit the data.     

\section{Experiments}

Experiments are conducted on the identical data and settings of \citet{ding2020three}, where identical ranks are assigned to words of same frequency, so that they are treated as one data point. In the calculation of $r_{\tt exp}$, all data points are used; those words with an identical rank are treated separately.\footnote{The results do not change much even if they are treated as one data point.}

Table \ref{europarl} listed the fitted data\footnote{The fitted $\beta$ of {\tt pl} is slightly different from the original paper that from $1.17$ to $1.18$. Similarly, the result on {\tt ro} is singular, where $t > 1/s$.} and figures are presented at the end of this manuscript. From the fitting experiments, there are the following observations.
\begin{itemize}
    \item The two parameters can generally fit the data as well as the four parameters do.
    \item Due to the introduction of the $r_{\tt max}$, the ending of the curve fits more exactly the rank of rare word.
    \item The two-parameter fitting is less sensitive to the rank of common words, which usually do not lie on a smooth curve. This can be considered as a tread-off by the reducing of the free parameters.
\end{itemize}

\begin{table}[t!]
\begin{center}
\begin{tabular}{|c|rr|rr|rrrr|}
\hline
& $\log_{10} r_{\tt max}$ & $\log_{10} r_{\tt exp}$ & $s$ & $t$ & $\alpha$ & $\beta$ & $\gamma$ & $C$ \\
\hline
\hline
{\tt bg} & $4.81$ & $3.01$ & $0.15$ & $4.80$ & $0.92$ & $2.05$ & $4.25$ & $14.59$ \\
{\tt cs} & $5.02$ & $3.58$ & $0.14$ & $3.04$ & $0.86$ & $1.20$ & $3.89$ & $10.56$ \\
{\tt da} & $5.26$ & $3.47$ & $0.02$ & $2.16$ & $0.99$ & $1.10$ & $3.85$ & $10.99$ \\
{\tt de} & $5.30$ & $3.55$ & $0.02$ & $2.14$ & $0.99$ & $1.08$ & $3.94$ & $11.00$ \\
{\tt el} & $5.14$ & $3.47$ & $0.13$ & $4.16$ & $0.98$ & $1.96$ & $4.43$ & $15.27$ \\
{\tt en} & $4.88$ & $3.01$ & $0.18$ & $4.74$ & $0.93$ & $2.04$ & $3.82$ & $14.52$ \\
{\tt es} & $5.05$ & $3.23$ & $0.14$ & $3.81$ & $0.94$ & $1.38$ & $3.82$ & $11.97$ \\
{\tt et} & $5.18$ & $3.85$ & $0.08$ & $3.58$ & $0.90$ & $1.06$ & $4.13$ & $10.23$ \\
{\tt fi} & $5.54$ & $4.10$ & $0.06$ & $3.21$ & $0.87$ & $0.89$ & $4.07$ & $9.88$ \\
{\tt fr} & $4.97$ & $3.13$ & $0.15$ & $5.69$ & $1.01$ & $2.05$ & $4.14$ & $15.37$ \\
{\tt hu} & $5.21$ & $3.85$ & $0.06$ & $4.19$ & $0.92$ & $0.96$ & $4.16$ & $9.90$ \\
{\tt it} & $5.03$ & $3.25$ & $0.18$ & $3.36$ & $0.94$ & $1.47$ & $3.84$ & $12.39$ \\
{\tt lt} & $5.09$ & $3.70$ & $0.10$ & $2.94$ & $0.84$ & $1.04$ & $3.77$ & $9.72$ \\
{\tt lv} & $4.99$ & $3.56$ & $0.16$ & $4.23$ & $0.87$ & $1.69$ & $4.22$ & $12.98$ \\
{\tt nl} & $5.16$ & $3.31$ & $0.06$ & $2.11$ & $0.98$ & $1.18$ & $3.73$ & $11.19$ \\
{\tt pl} & $5.04$ & $3.61$ & $0.14$ & $3.19$ & $0.87$ & $1.17$ & $3.97$ & $10.56$ \\
{\tt pt} & $5.05$ & $3.26$ & $0.16$ & $3.40$ & $0.93$ & $1.33$ & $3.77$ & $11.70$ \\
{\tt ro} & $4.74$ & $3.29$ & $0.17$ & $10.08$ & $0.94$ & $5.24$ & $4.78$ & $30.93$ \\
{\tt sk} & $5.03$ & $3.60$ & $0.14$ & $3.45$ & $0.89$ & $1.38$ & $4.14$ & $11.68$ \\
{\tt sl} & $4.96$ & $3.50$ & $0.16$ & $4.78$ & $0.91$ & $1.77$ & $4.31$ & $13.57$ \\
{\tt sv} & $5.26$ & $3.51$ & $0.03$ & $2.13$ & $0.99$ & $1.05$ & $3.86$ & $10.76$ \\
\hline
\end{tabular}
\end{center}
\caption{Fitted parameters on Europarl data.}
\label{europarl}
\end{table}

\section{Conclusion}

This manuscript provides a two-parameter formulation on word rank-frequency relations. The formulation has a general fitting capacity as a previous four-parameter formulation.

\bibliographystyle{plainnat}
\bibliography{ref}

\newpage

\begin{figure}[t]
\centering
\begin{minipage}{0.45\linewidth}
    \centering
    \includegraphics[width=1.0\linewidth]{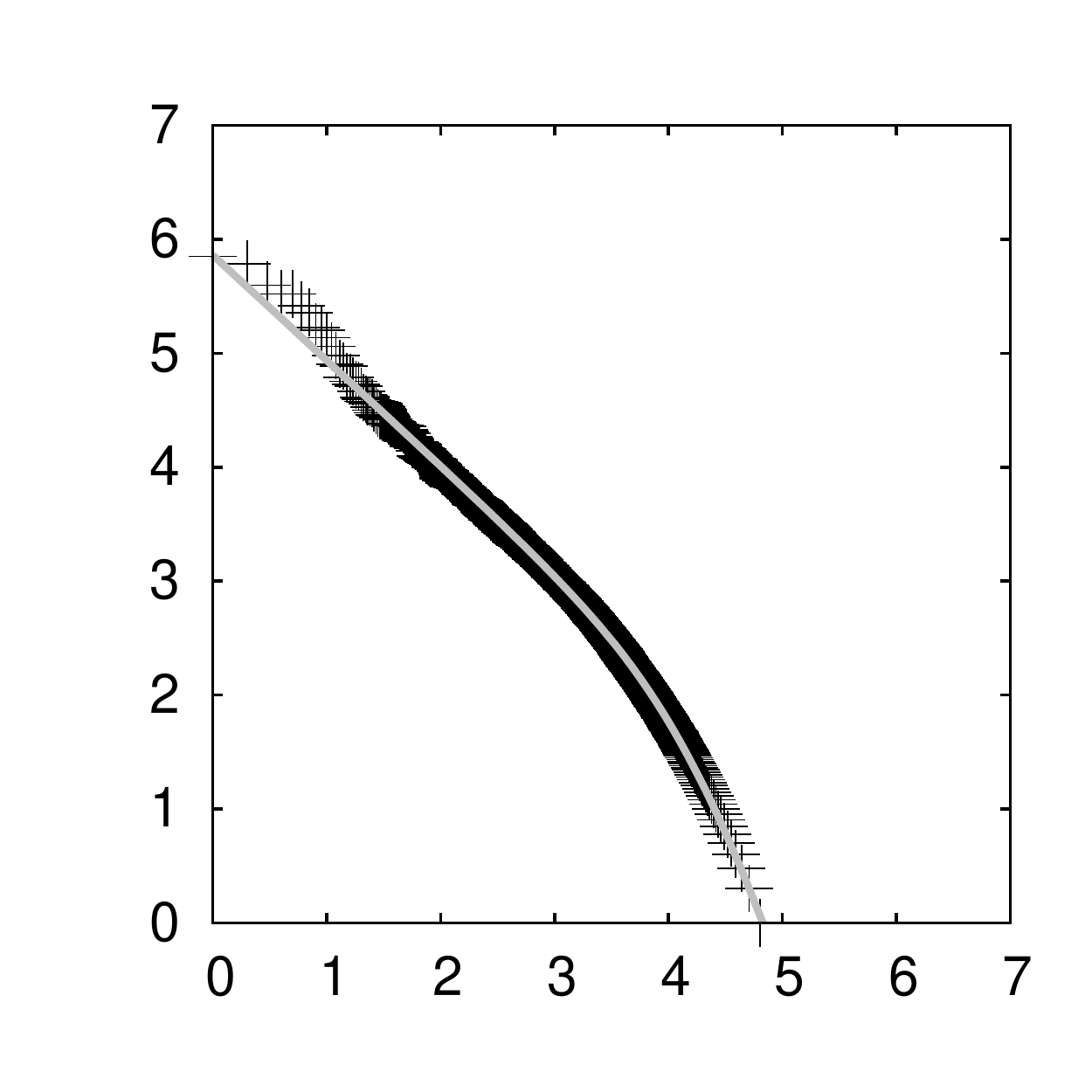}
\end{minipage}
\begin{minipage}{0.45\linewidth}
    \centering
    \includegraphics[width=1.0\linewidth]{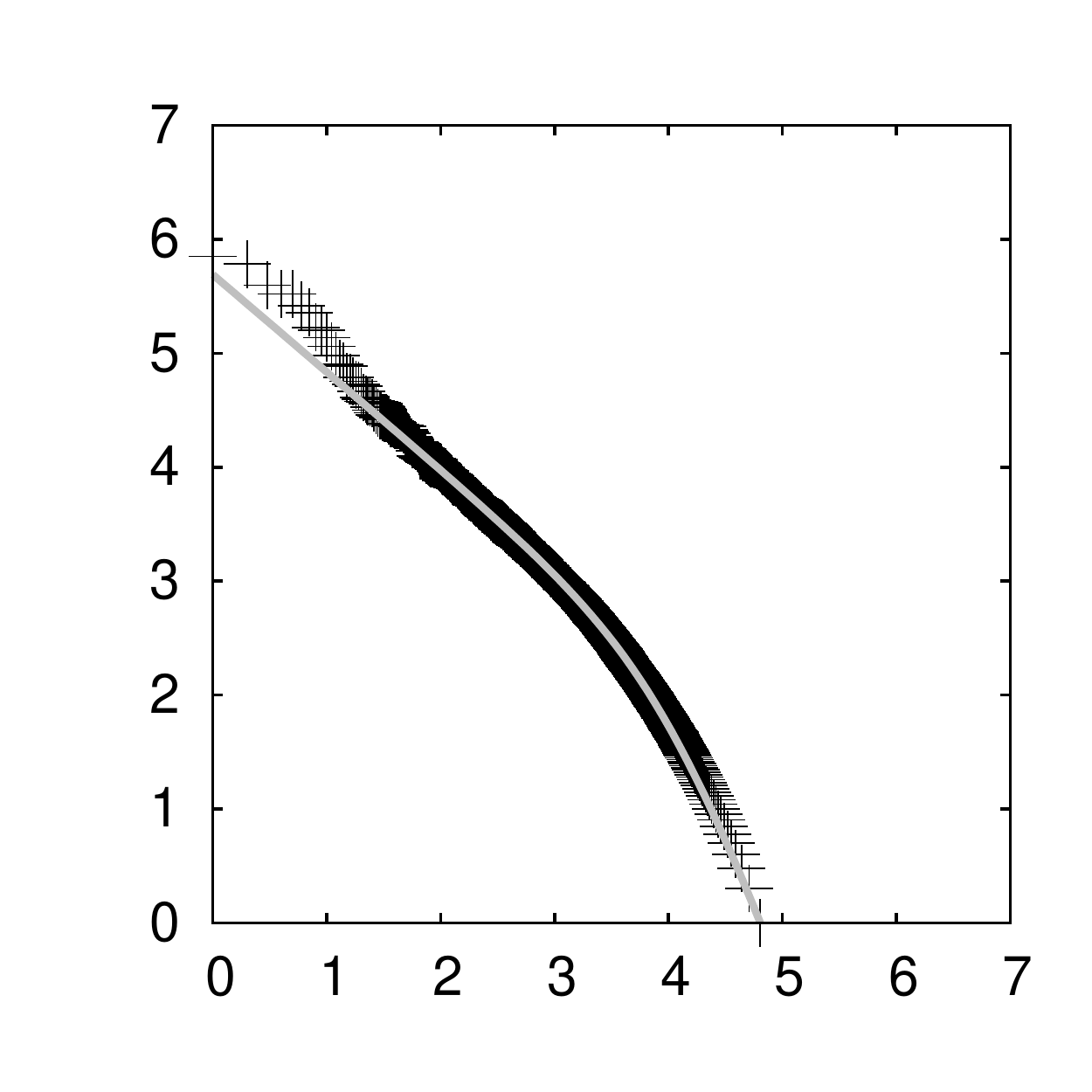}
\end{minipage}
\vspace{-3mm}
\caption{four-parameter (left) and two-parameter (right) fitting on {\tt bg}.}
\vspace{-3mm}
\label{fig:bg}
\end{figure}

\begin{figure}[t]
\centering
\begin{minipage}{0.45\linewidth}
    \centering
    \includegraphics[width=1.0\linewidth]{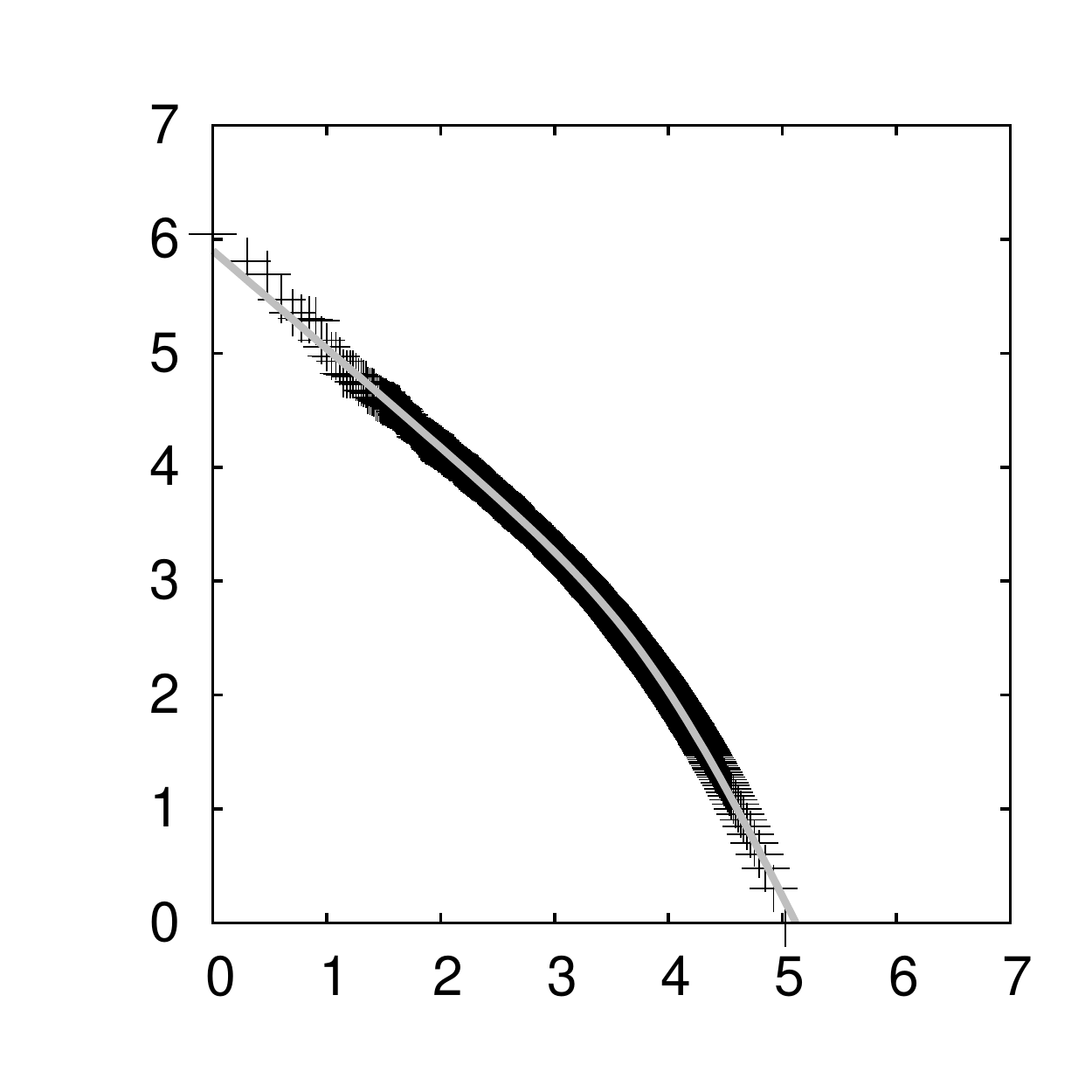}
\end{minipage}
\begin{minipage}{0.45\linewidth}
    \centering
    \includegraphics[width=1.0\linewidth]{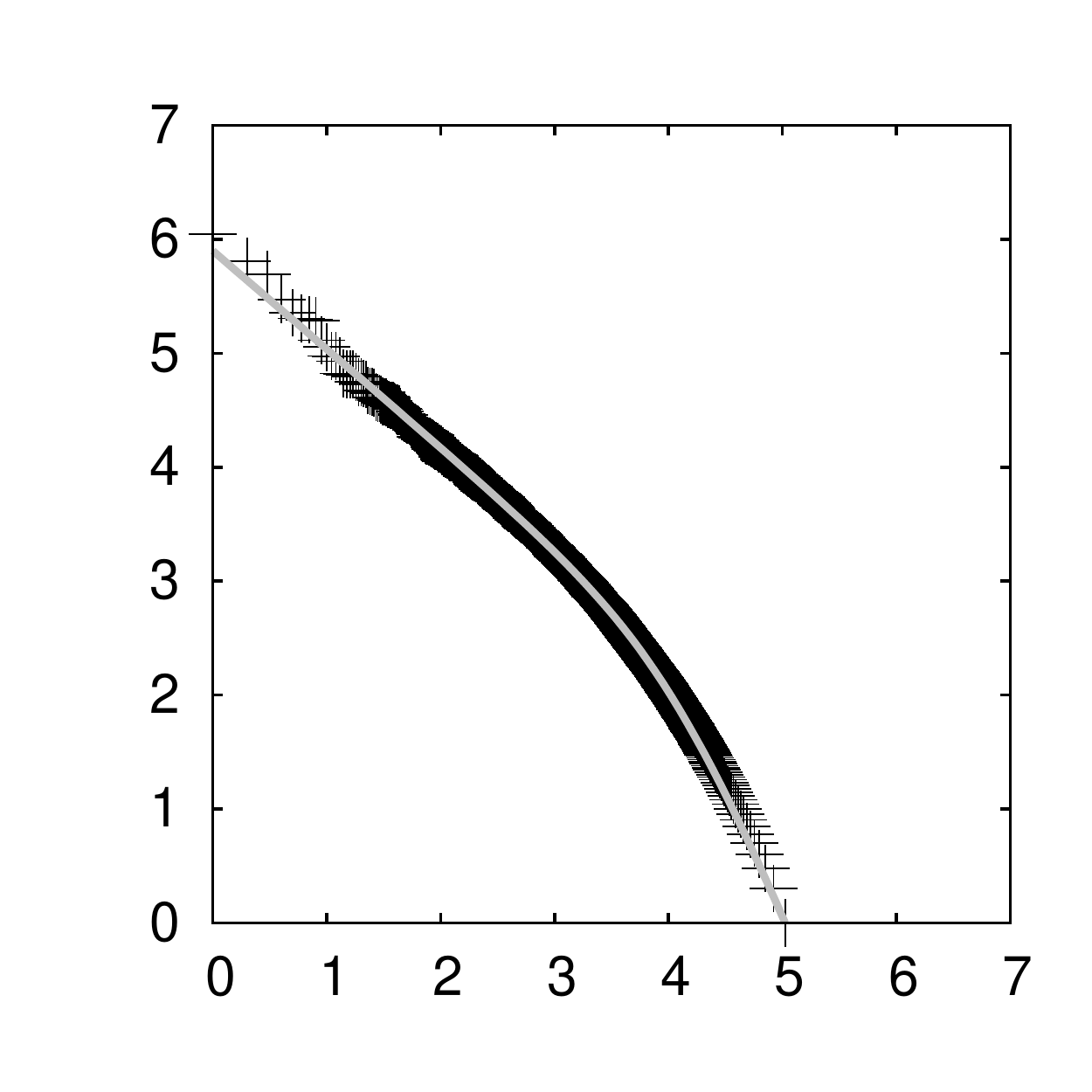}
\end{minipage}
\vspace{-3mm}
\caption{four-parameter (left) and two-parameter (right) fitting on {\tt cs}.}
\vspace{-3mm}
\label{fig:cs}
\end{figure}

\begin{figure}[t]
\centering
\begin{minipage}{0.45\linewidth}
    \centering
    \includegraphics[width=1.0\linewidth]{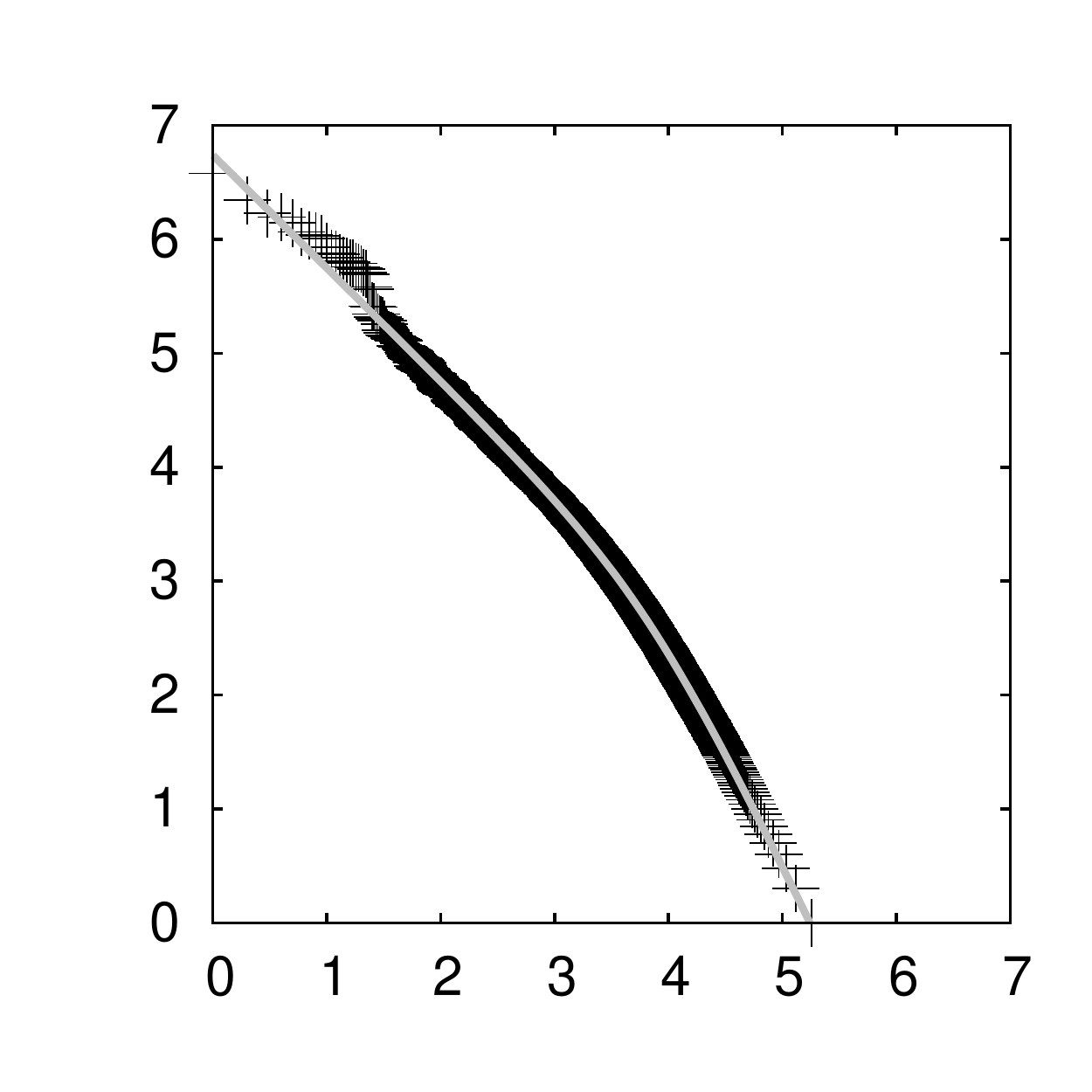}
\end{minipage}
\begin{minipage}{0.45\linewidth}
    \centering
    \includegraphics[width=1.0\linewidth]{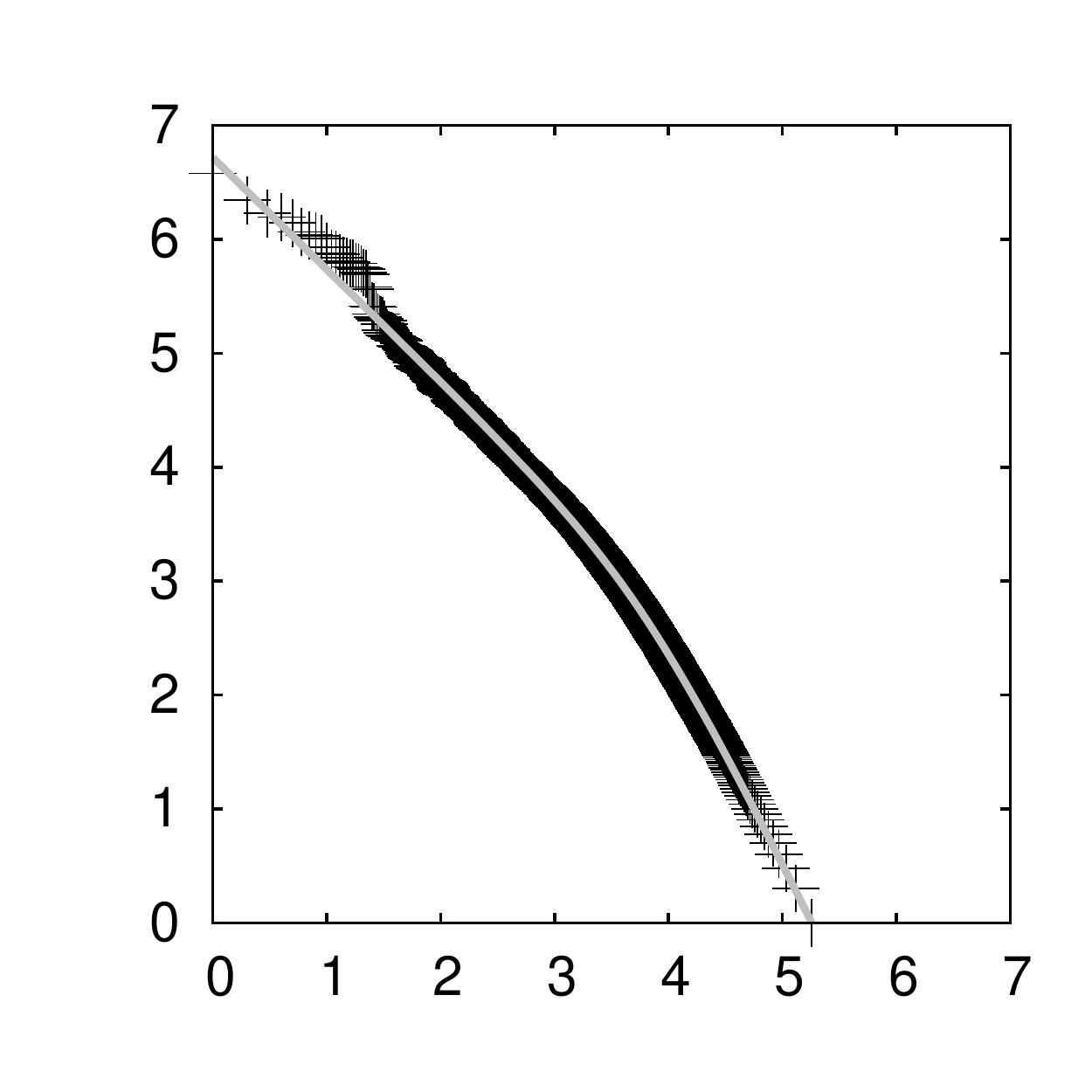}
\end{minipage}
\vspace{-3mm}
\caption{four-parameter (left) and two-parameter (right) fitting on {\tt da}.}
\vspace{-3mm}
\label{fig:da}
\end{figure}

\begin{figure}[t]
\centering
\begin{minipage}{0.45\linewidth}
    \centering
    \includegraphics[width=1.0\linewidth]{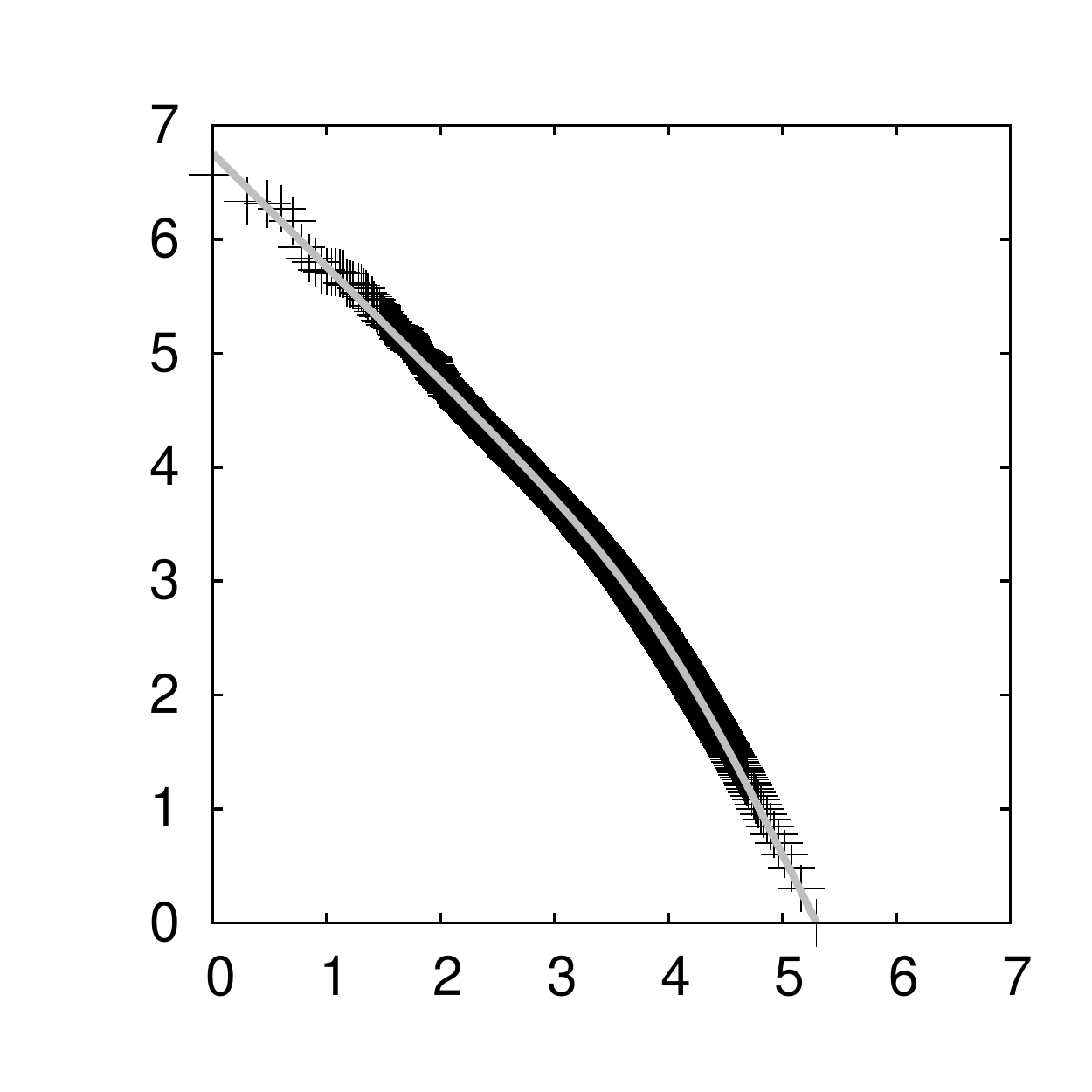}
\end{minipage}
\begin{minipage}{0.45\linewidth}
    \centering
    \includegraphics[width=1.0\linewidth]{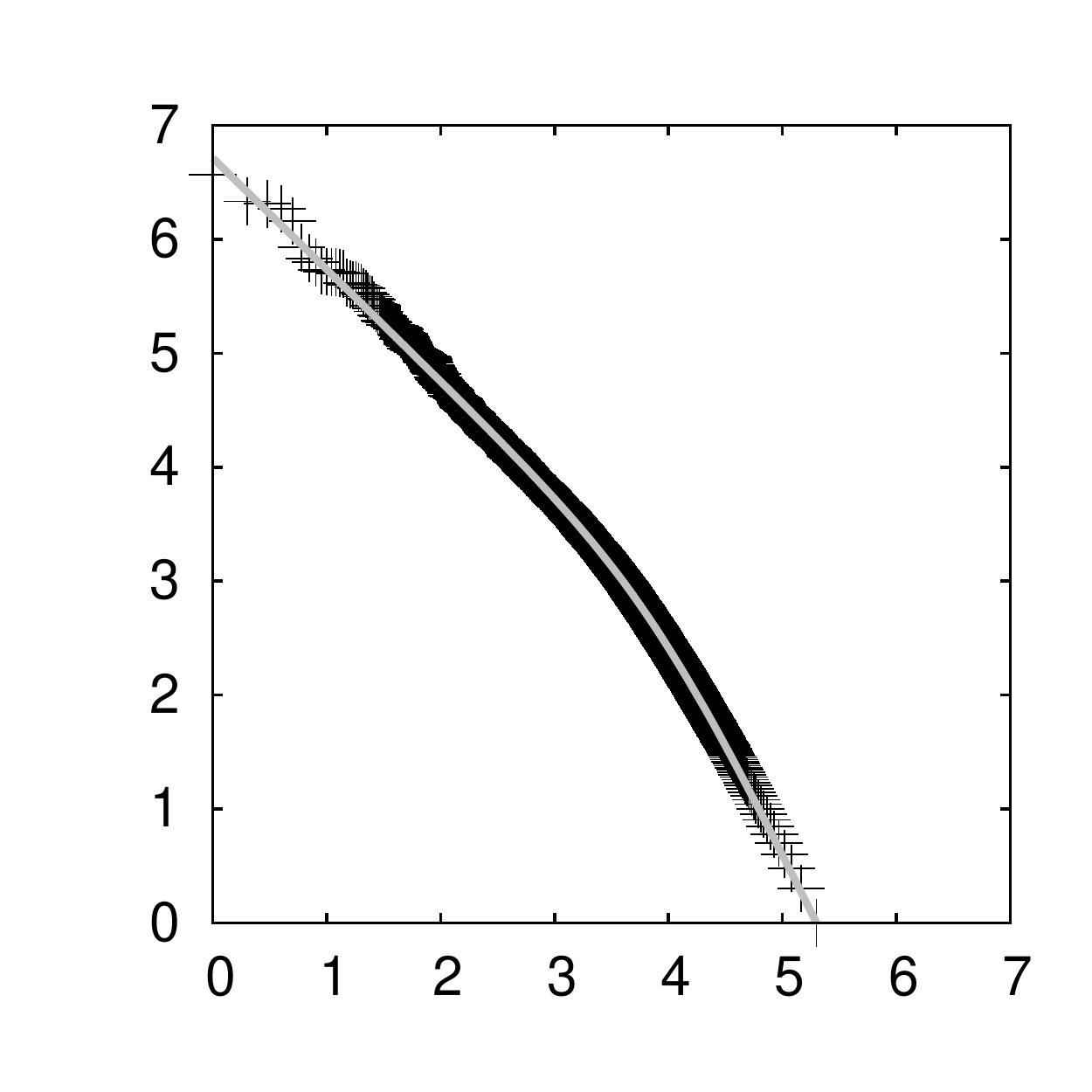}
\end{minipage}
\vspace{-3mm}
\caption{four-parameter (left) and two-parameter (right) fitting on {\tt de}.}
\vspace{-3mm}
\label{fig:de}
\end{figure}

\begin{figure}[t]
\centering
\begin{minipage}{0.45\linewidth}
    \centering
    \includegraphics[width=1.0\linewidth]{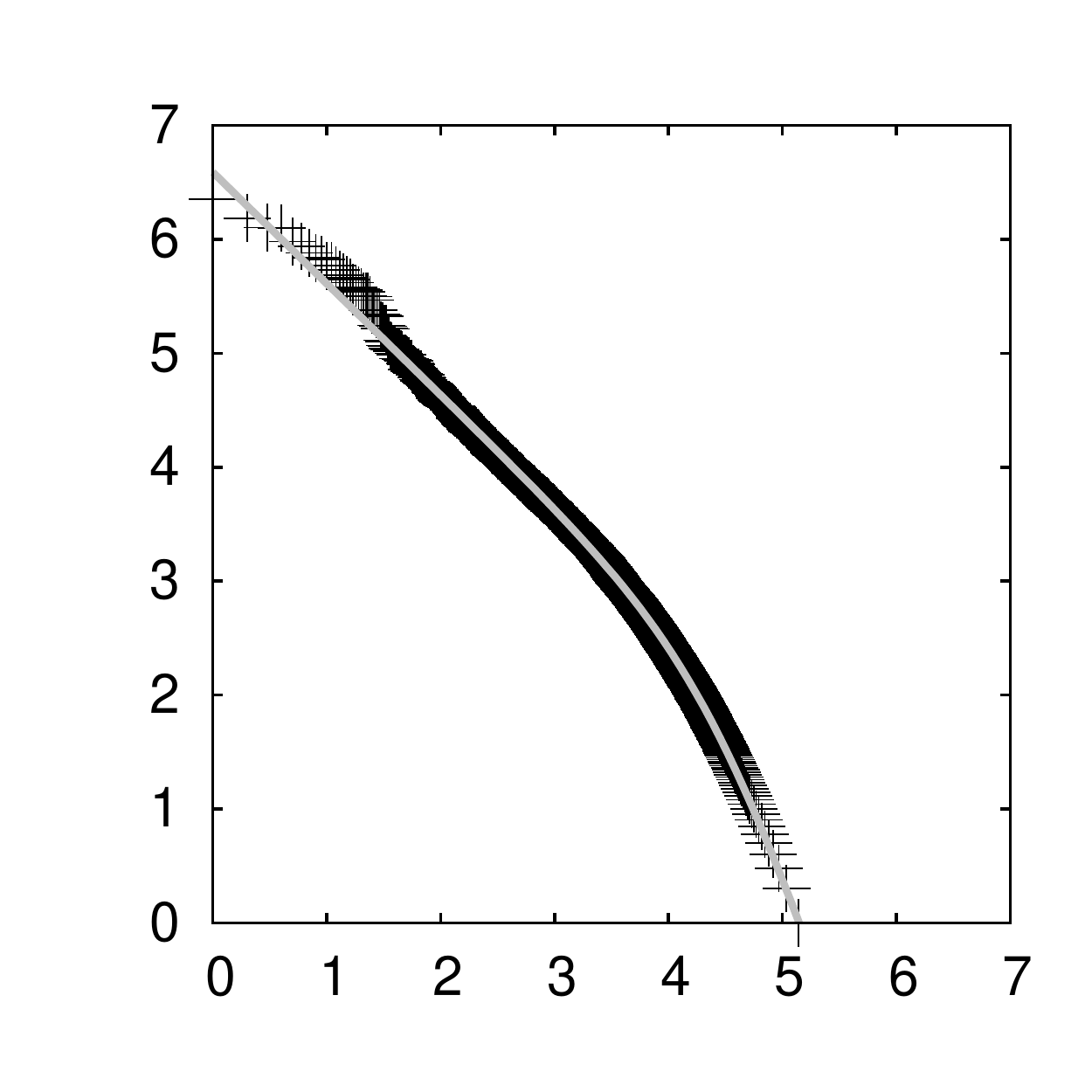}
\end{minipage}
\begin{minipage}{0.45\linewidth}
    \centering
    \includegraphics[width=1.0\linewidth]{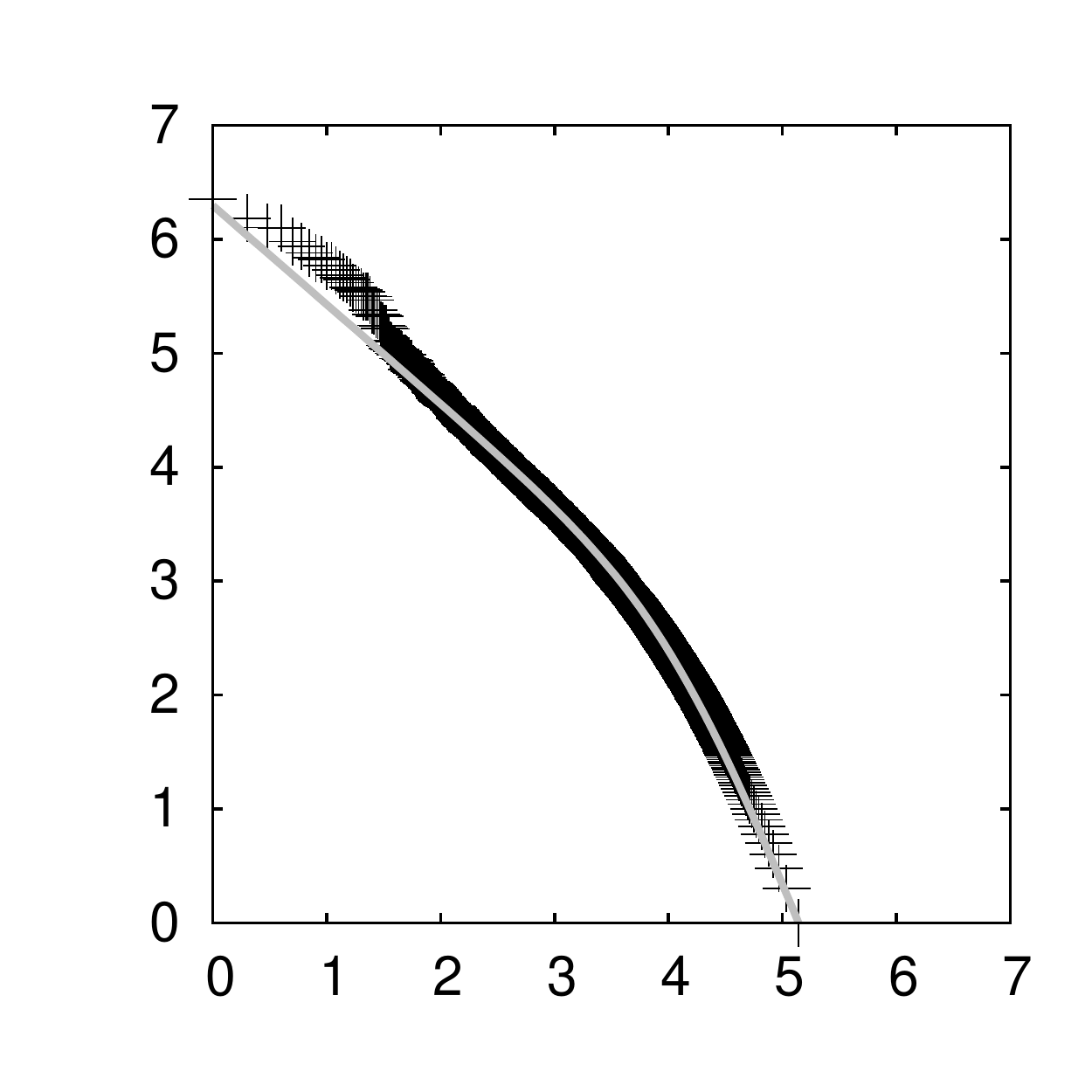}
\end{minipage}
\vspace{-3mm}
\caption{four-parameter (left) and two-parameter (right) fitting on {\tt el}.}
\vspace{-3mm}
\label{fig:el}
\end{figure}

\begin{figure}[t]
\centering
\begin{minipage}{0.45\linewidth}
    \centering
    \includegraphics[width=1.0\linewidth]{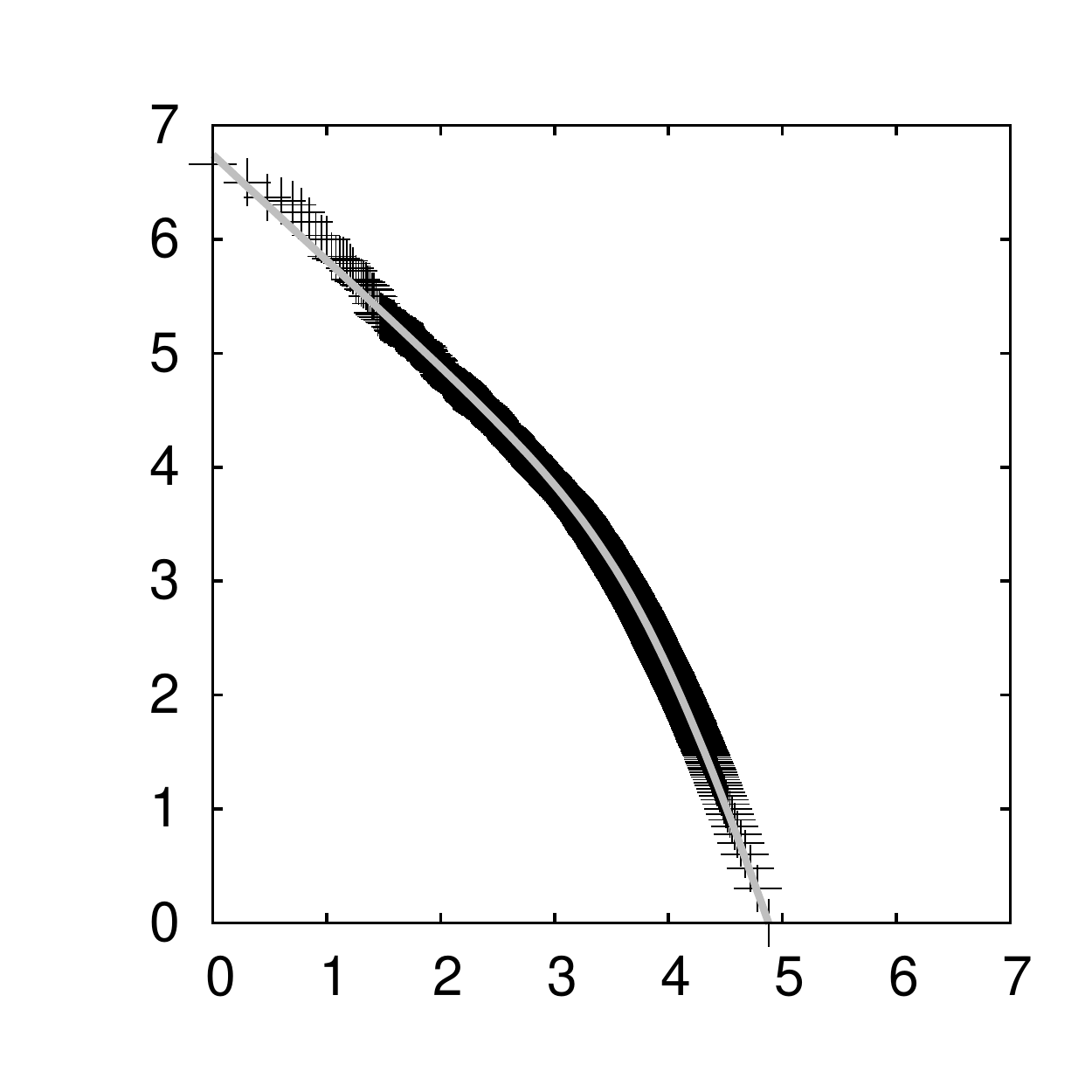}
\end{minipage}
\begin{minipage}{0.45\linewidth}
    \centering
    \includegraphics[width=1.0\linewidth]{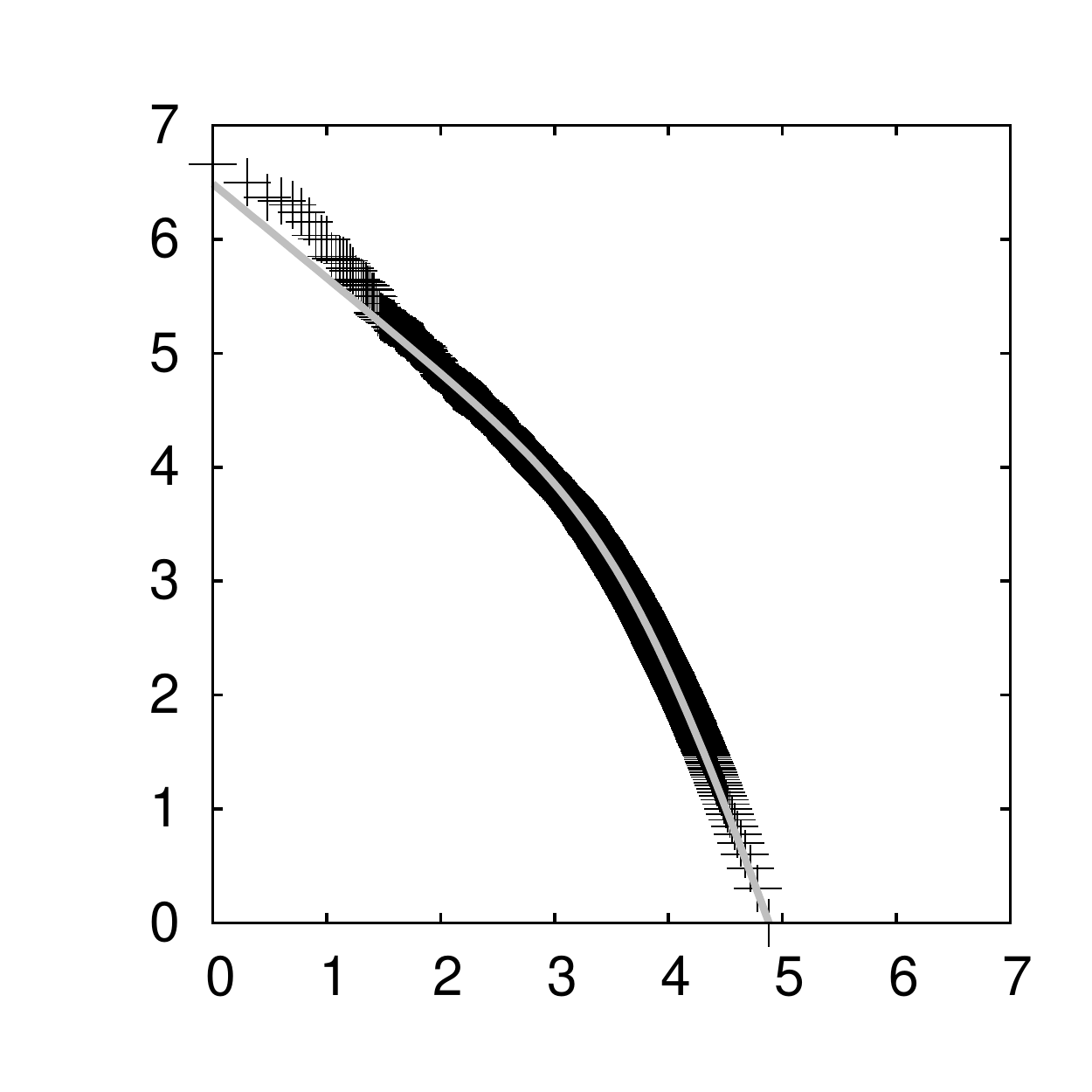}
\end{minipage}
\vspace{-3mm}
\caption{four-parameter (left) and two-parameter (right) fitting on {\tt en}.}
\vspace{-3mm}
\label{fig:en}
\end{figure}

\begin{figure}[t]
\centering
\begin{minipage}{0.45\linewidth}
    \centering
    \includegraphics[width=1.0\linewidth]{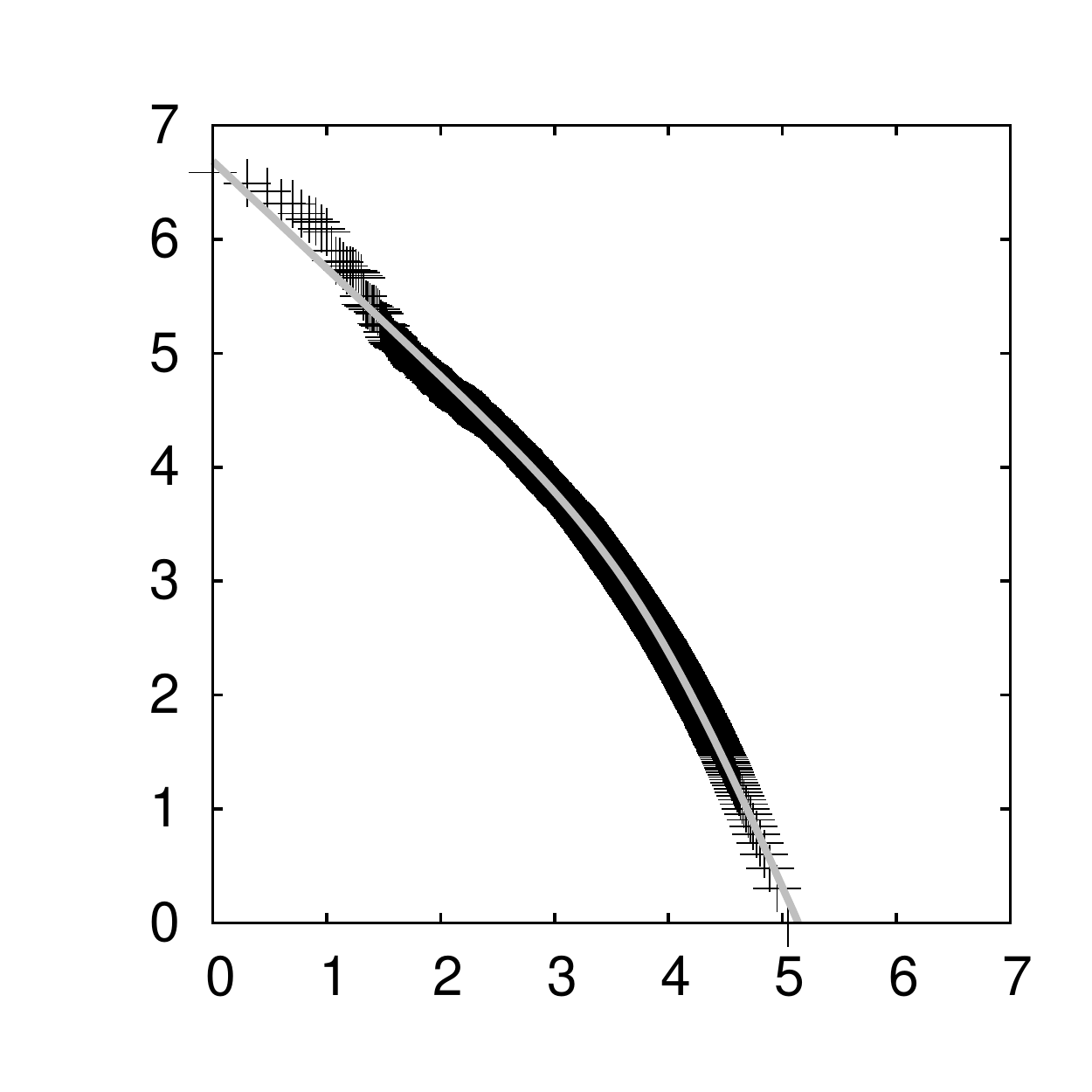}
\end{minipage}
\begin{minipage}{0.45\linewidth}
    \centering
    \includegraphics[width=1.0\linewidth]{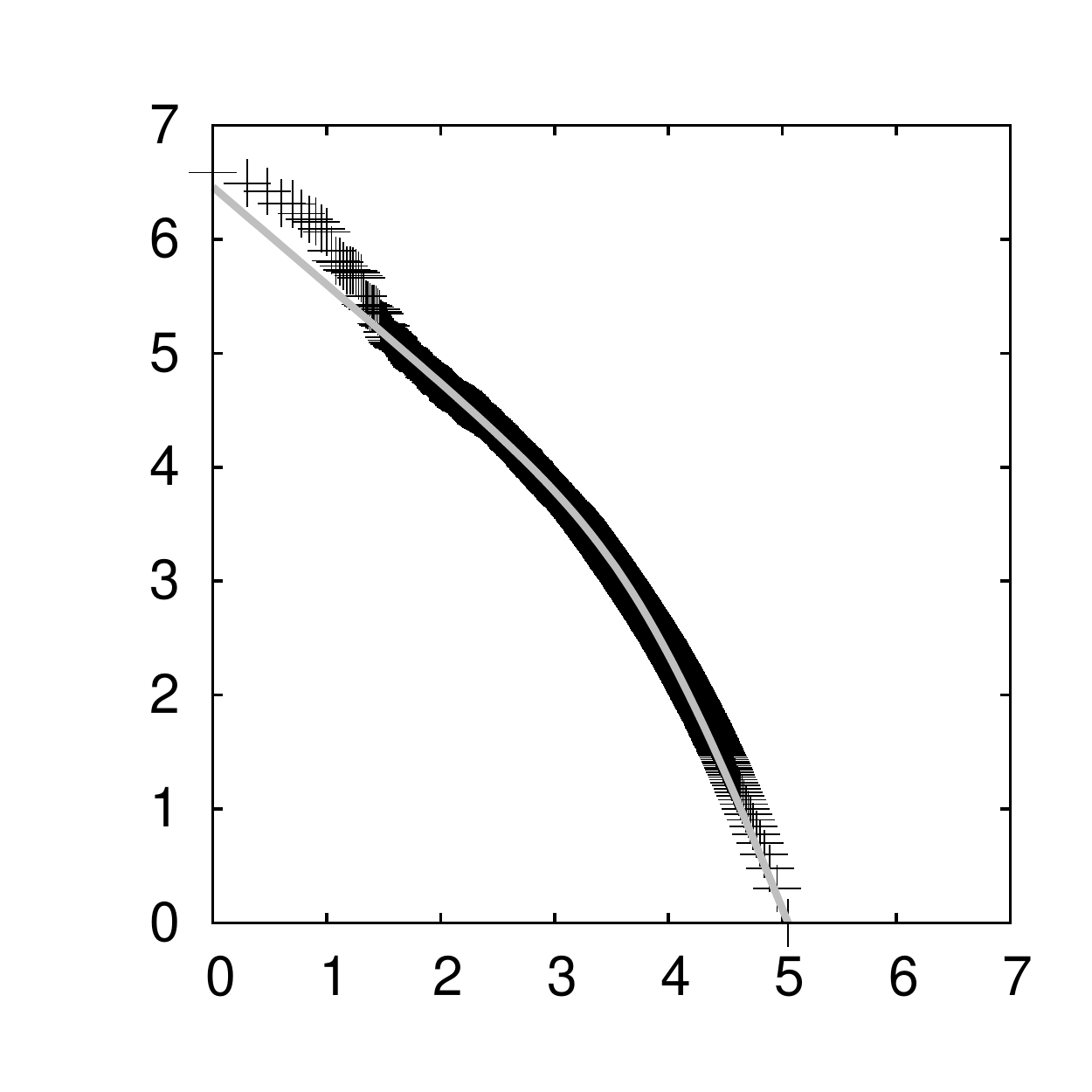}
\end{minipage}
\vspace{-3mm}
\caption{four-parameter (left) and two-parameter (right) fitting on {\tt es}.}
\vspace{-3mm}
\label{fig:es}
\end{figure}

\begin{figure}[t]
\centering
\begin{minipage}{0.45\linewidth}
    \centering
    \includegraphics[width=1.0\linewidth]{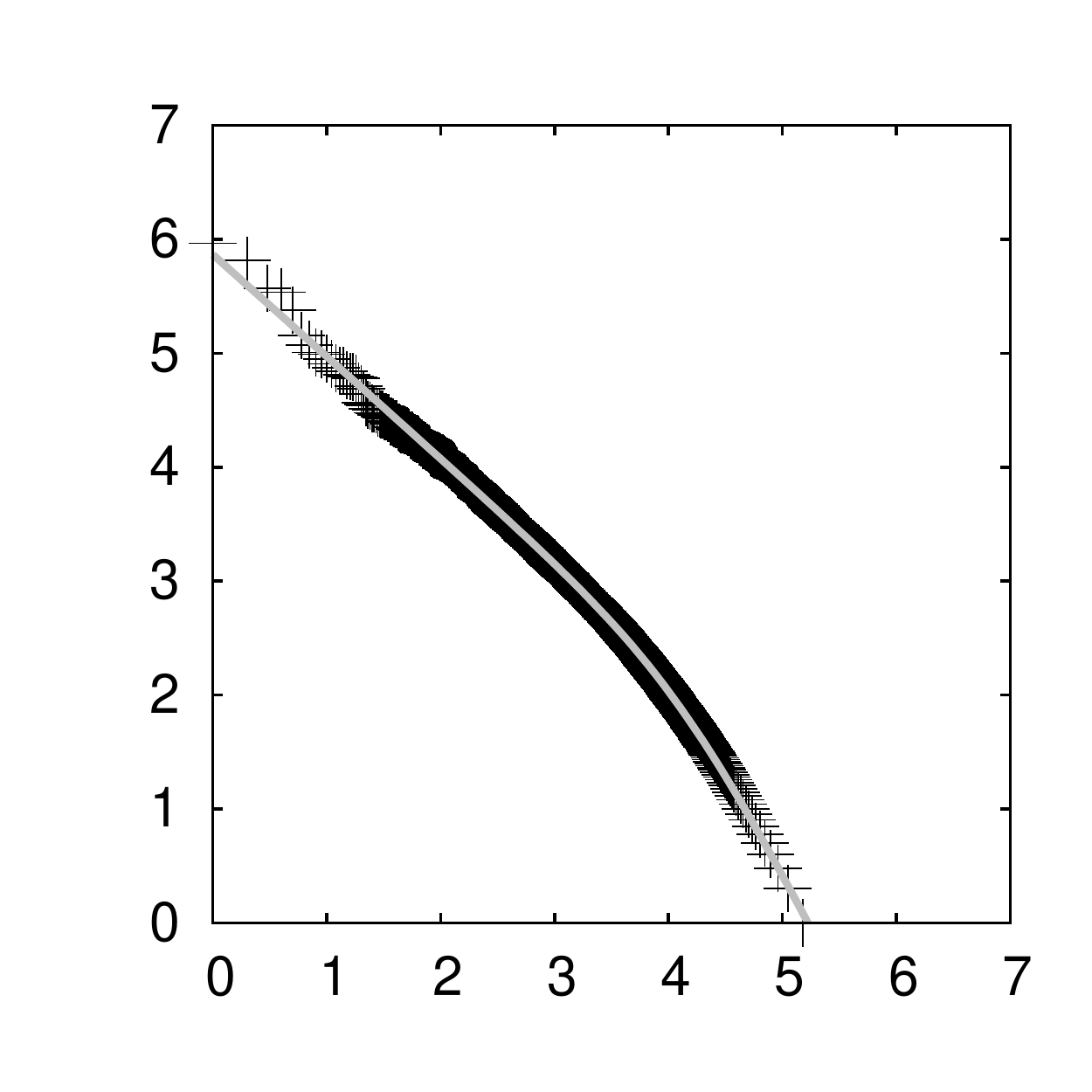}
\end{minipage}
\begin{minipage}{0.45\linewidth}
    \centering
    \includegraphics[width=1.0\linewidth]{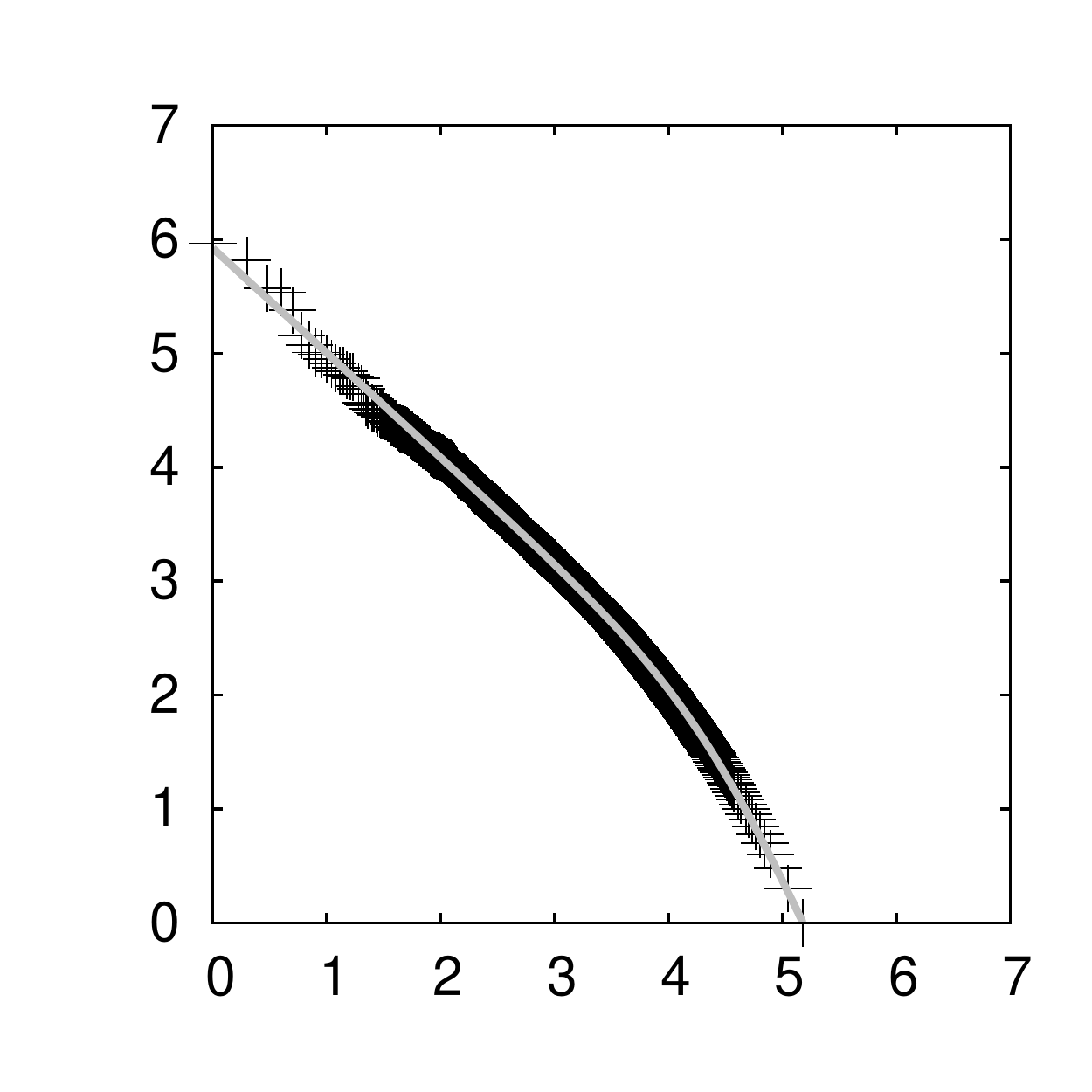}
\end{minipage}
\vspace{-3mm}
\caption{four-parameter (left) and two-parameter (right) fitting on {\tt et}.}
\vspace{-3mm}
\label{fig:et}
\end{figure}

\begin{figure}[t]
\centering
\begin{minipage}{0.45\linewidth}
    \centering
    \includegraphics[width=1.0\linewidth]{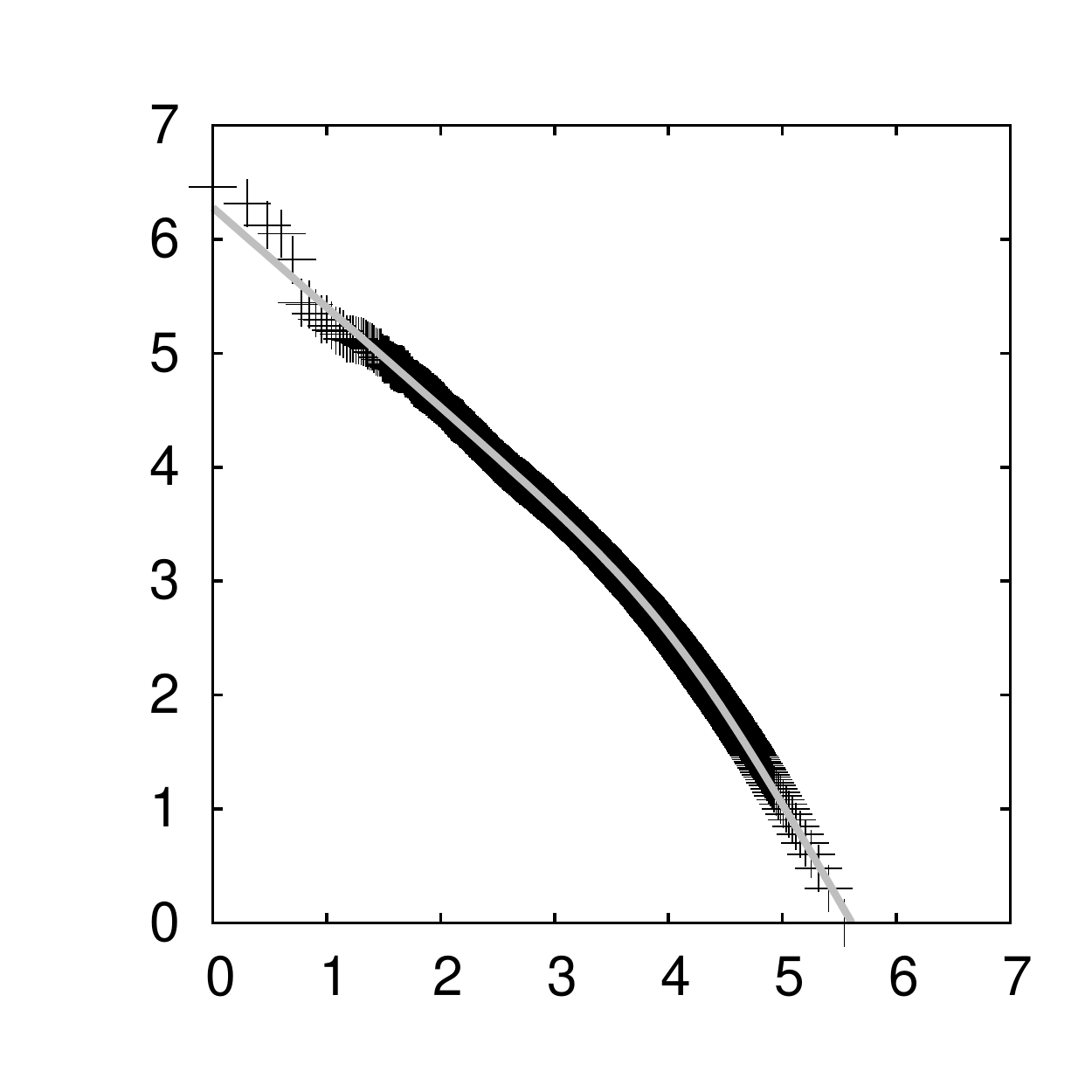}
\end{minipage}
\begin{minipage}{0.45\linewidth}
    \centering
    \includegraphics[width=1.0\linewidth]{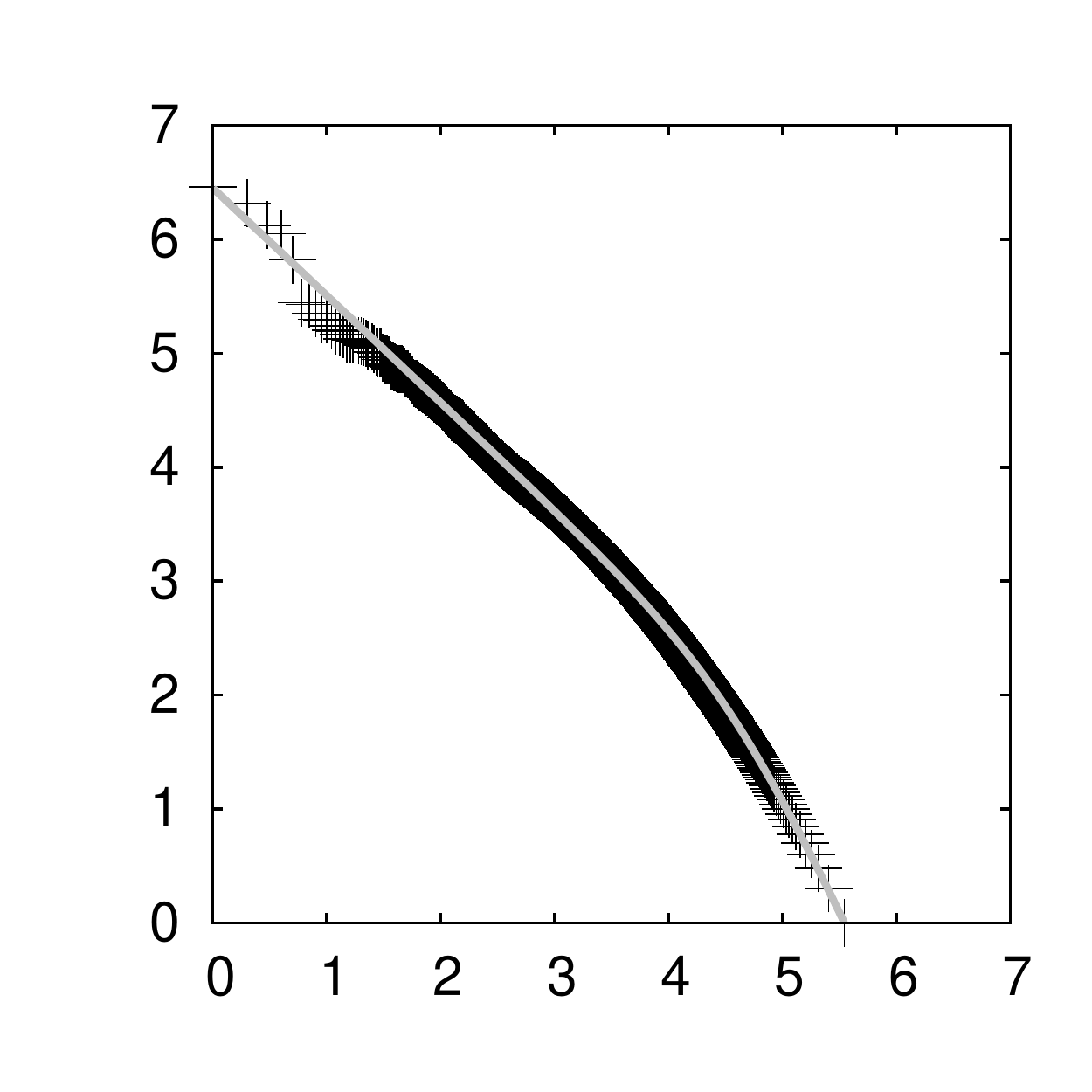}
\end{minipage}
\vspace{-3mm}
\caption{four-parameter (left) and two-parameter (right) fitting on {\tt fi}.}
\vspace{-3mm}
\label{fig:fi}
\end{figure}

\begin{figure}[t]
\centering
\begin{minipage}{0.45\linewidth}
    \centering
    \includegraphics[width=1.0\linewidth]{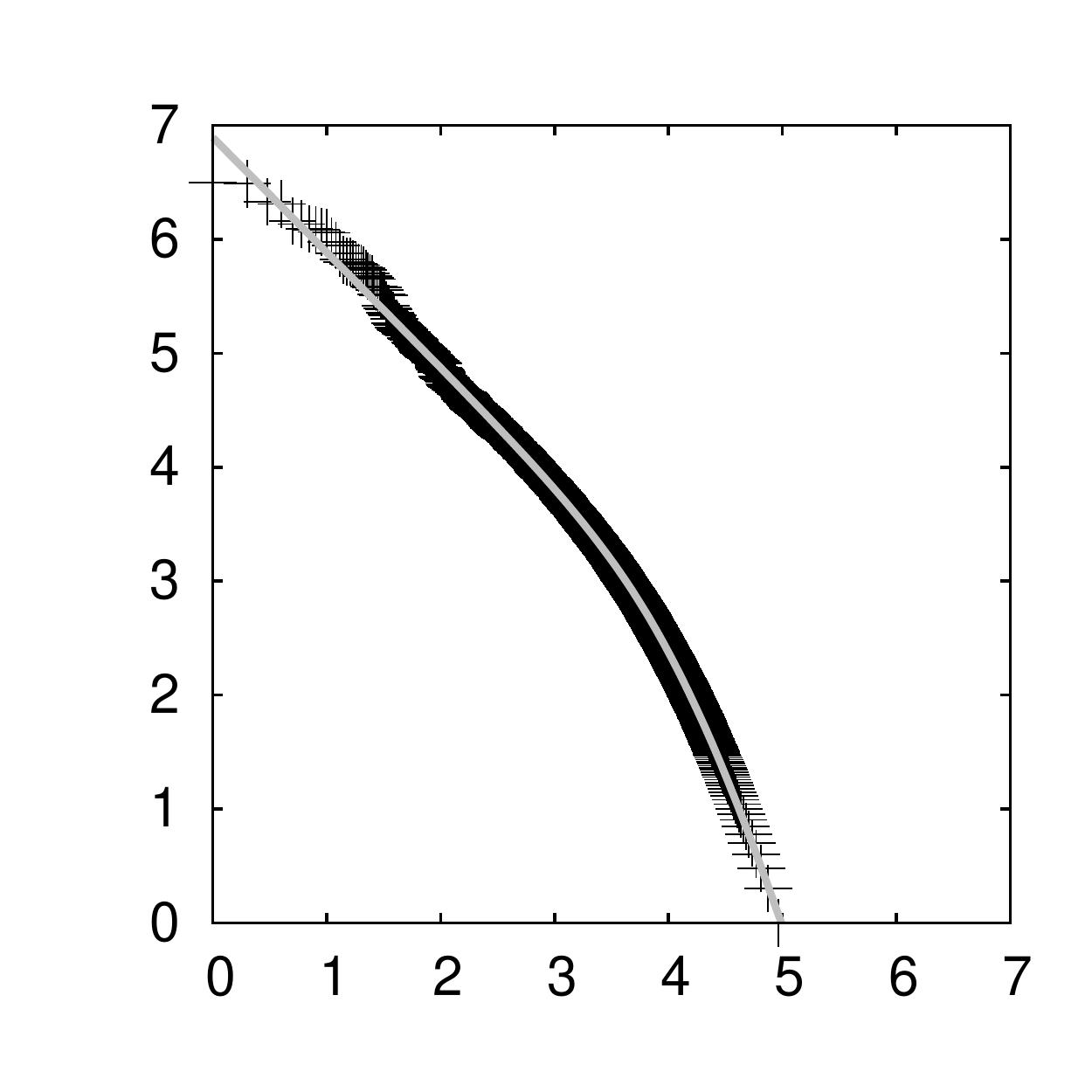}
\end{minipage}
\begin{minipage}{0.45\linewidth}
    \centering
    \includegraphics[width=1.0\linewidth]{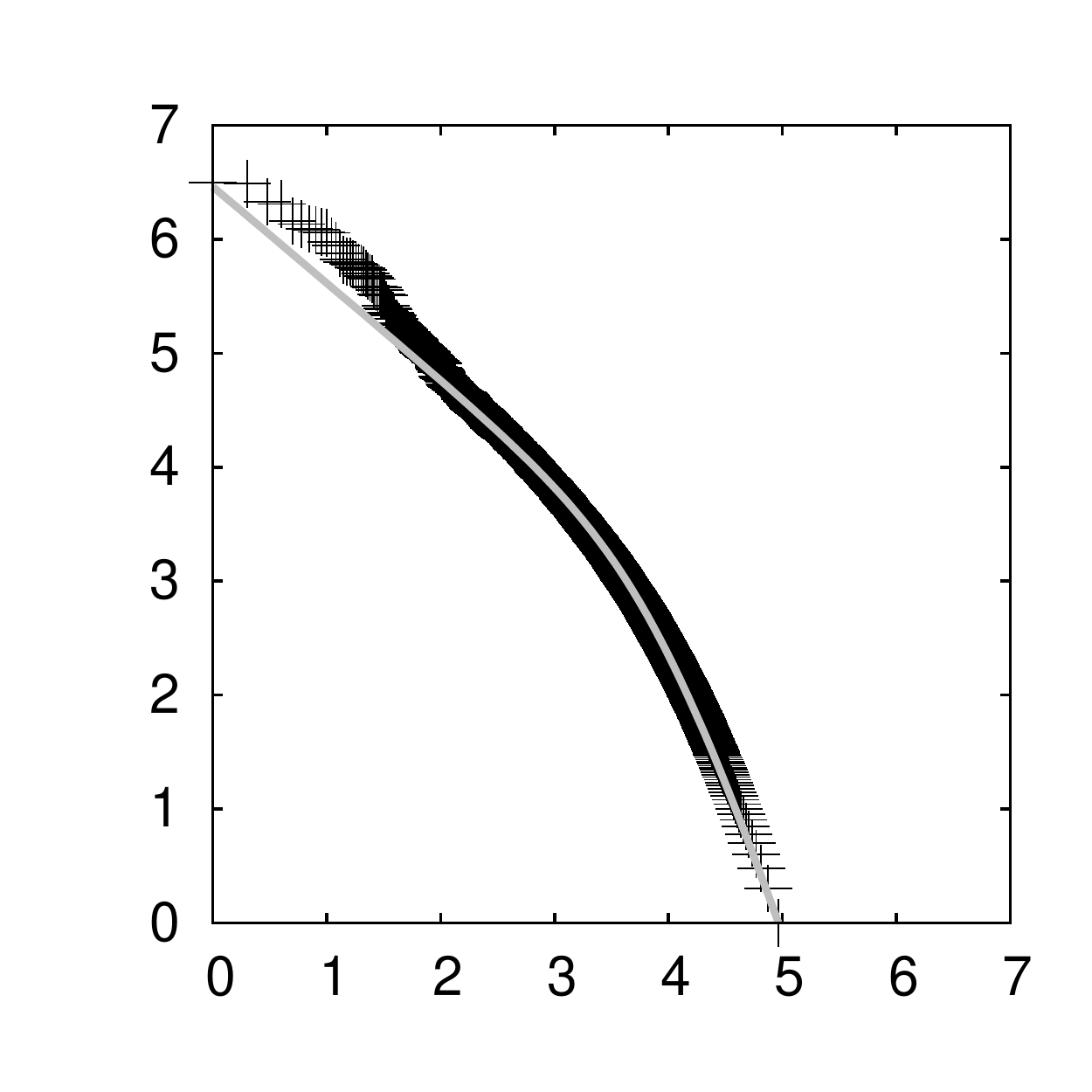}
\end{minipage}
\vspace{-3mm}
\caption{four-parameter (left) and two-parameter (right) fitting on {\tt fr}.}
\vspace{-3mm}
\label{fig:fr}
\end{figure}

\begin{figure}[t]
\centering
\begin{minipage}{0.45\linewidth}
    \centering
    \includegraphics[width=1.0\linewidth]{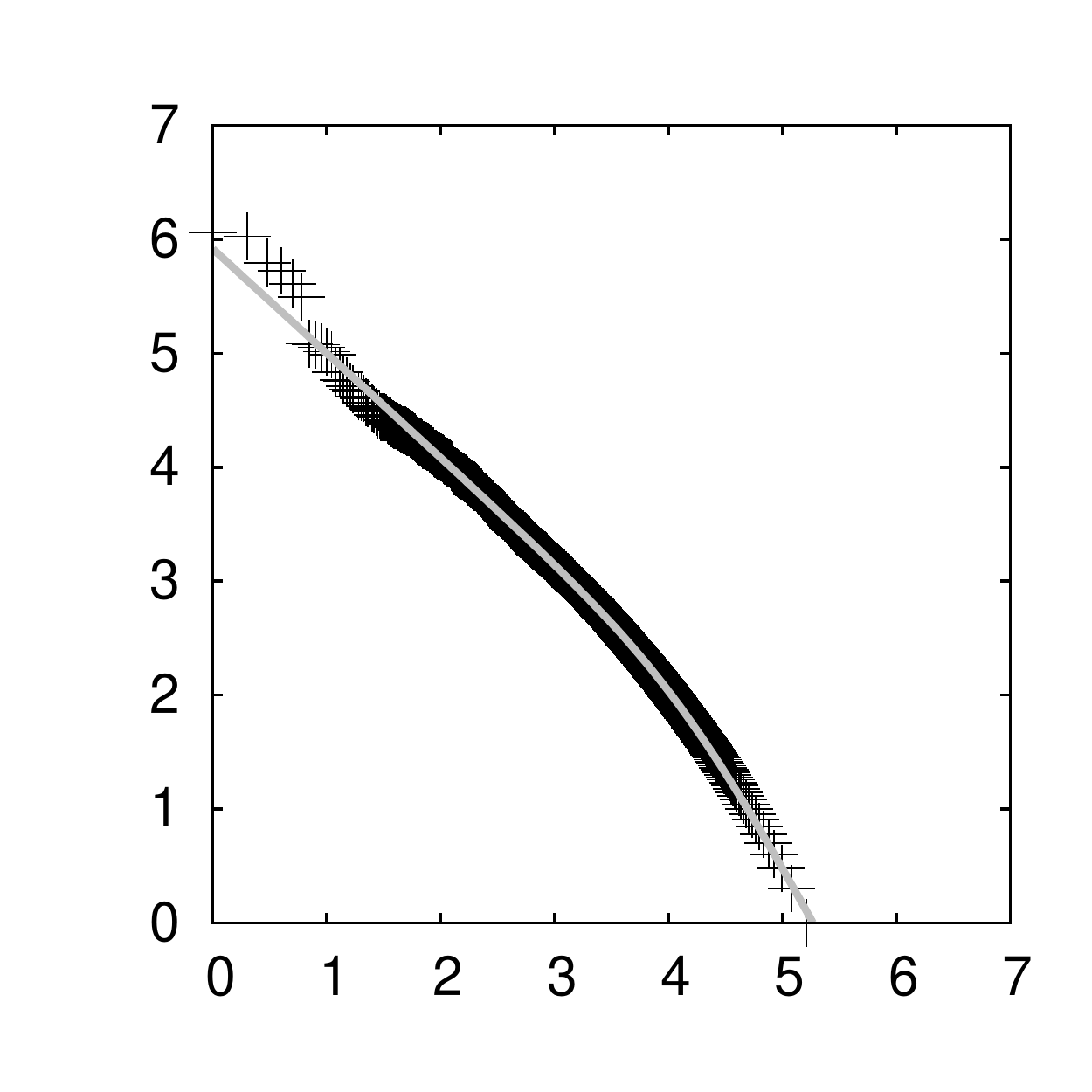}
\end{minipage}
\begin{minipage}{0.45\linewidth}
    \centering
    \includegraphics[width=1.0\linewidth]{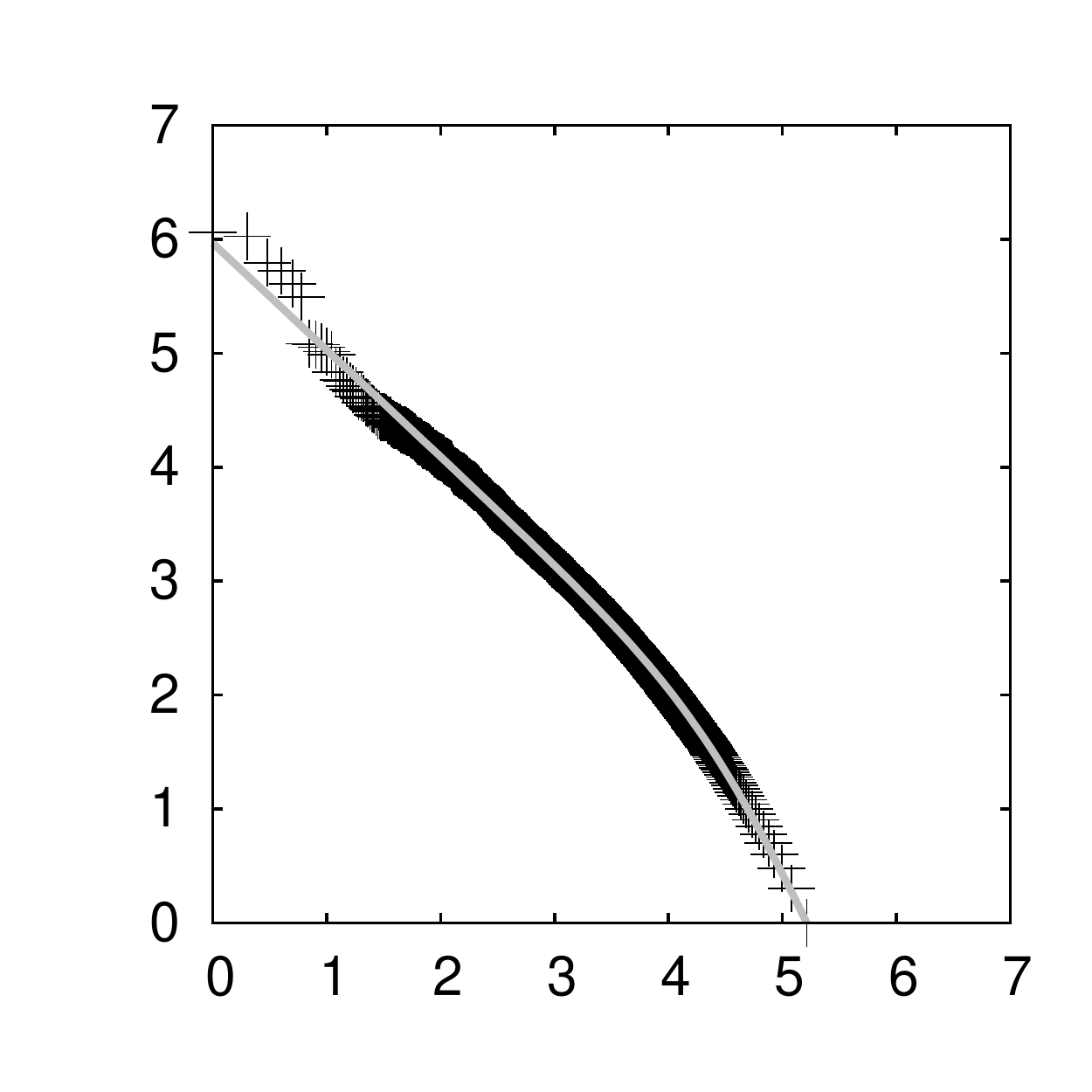}
\end{minipage}
\vspace{-3mm}
\caption{four-parameter (left) and two-parameter (right) fitting on {\tt hu}.}
\vspace{-3mm}
\label{fig:hu}
\end{figure}

\begin{figure}[t]
\centering
\begin{minipage}{0.45\linewidth}
    \centering
    \includegraphics[width=1.0\linewidth]{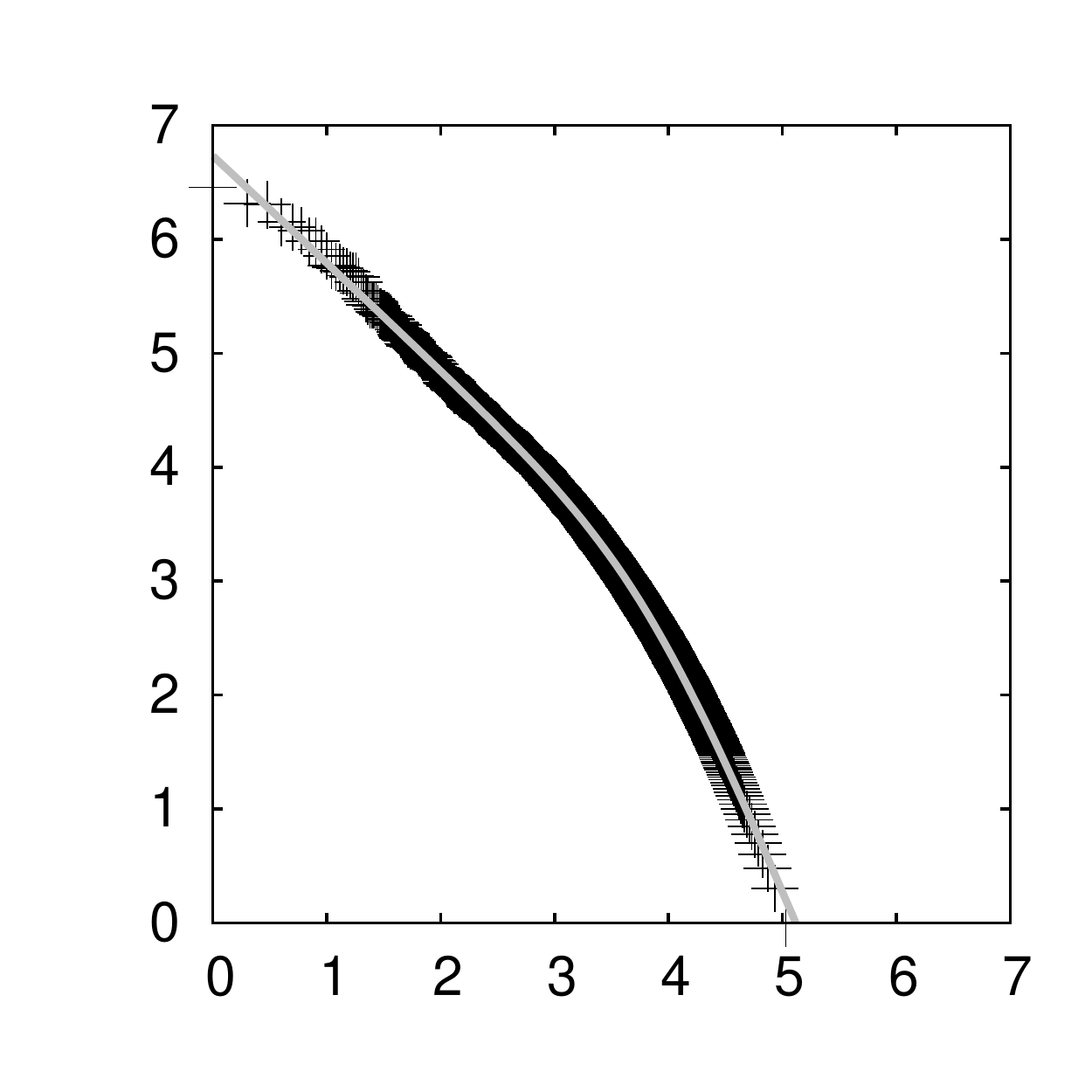}
\end{minipage}
\begin{minipage}{0.45\linewidth}
    \centering
    \includegraphics[width=1.0\linewidth]{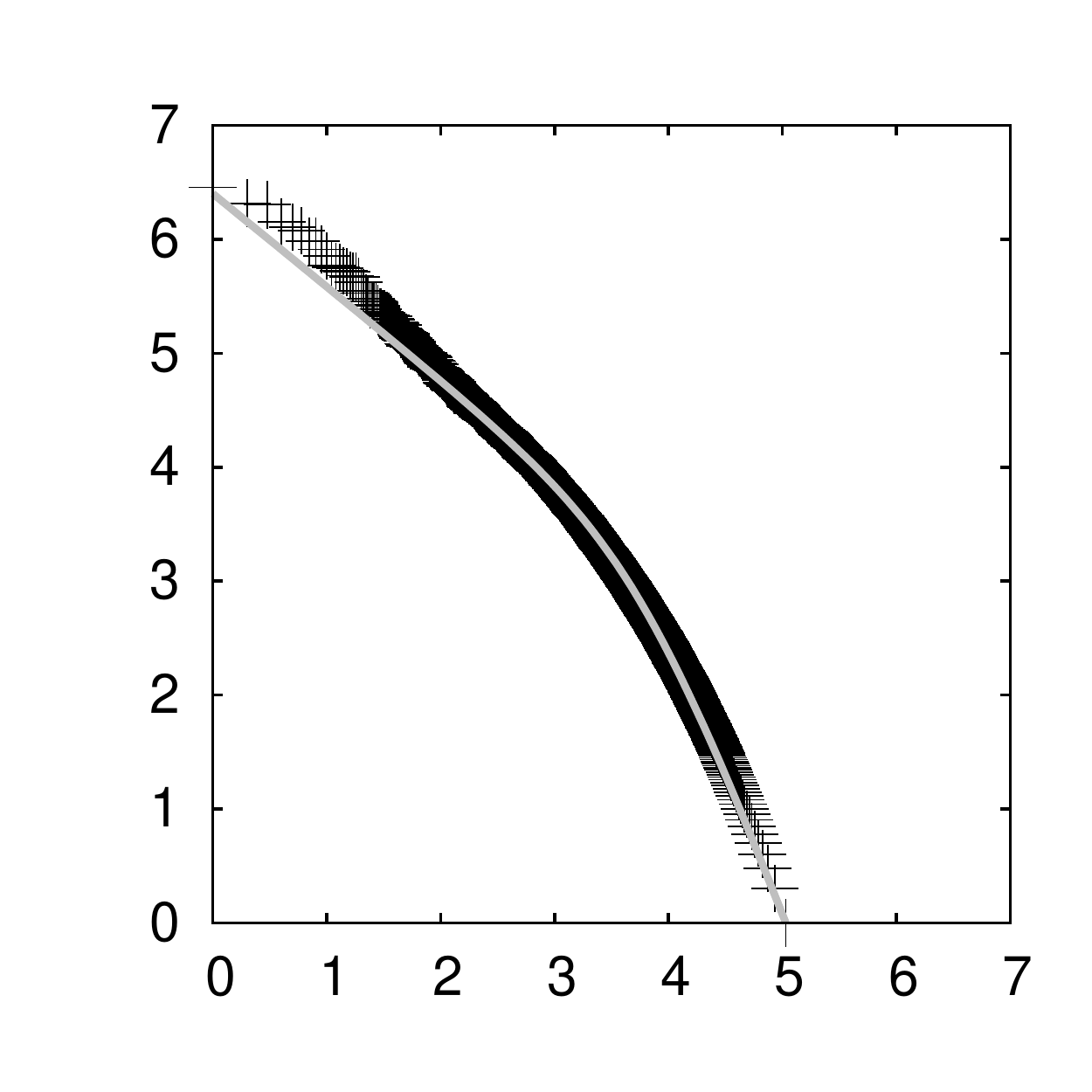}
\end{minipage}
\vspace{-3mm}
\caption{four-parameter (left) and two-parameter (right) fitting on {\tt it}.}
\vspace{-3mm}
\label{fig:it}
\end{figure}

\begin{figure}[t]
\centering
\begin{minipage}{0.45\linewidth}
    \centering
    \includegraphics[width=1.0\linewidth]{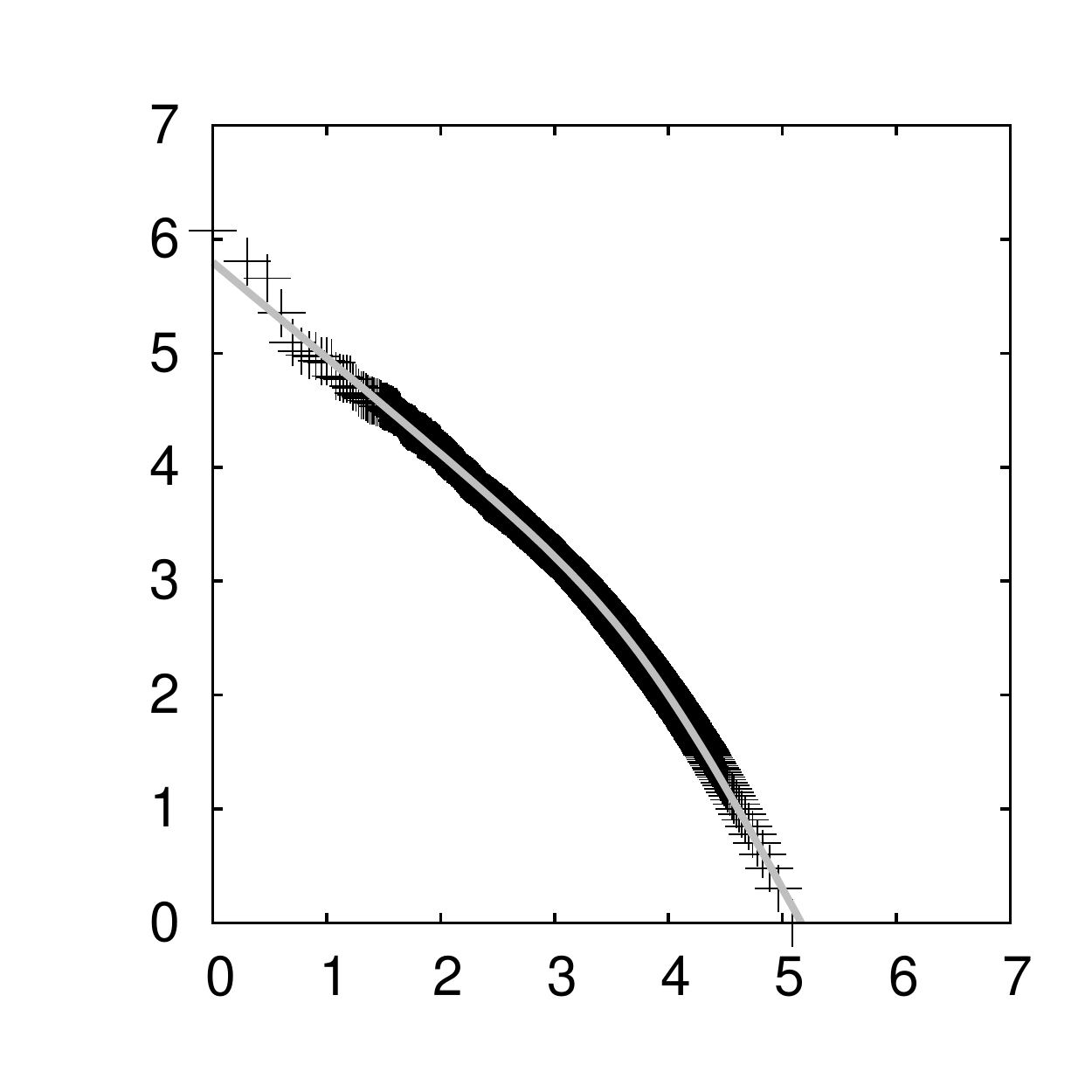}
\end{minipage}
\begin{minipage}{0.45\linewidth}
    \centering
    \includegraphics[width=1.0\linewidth]{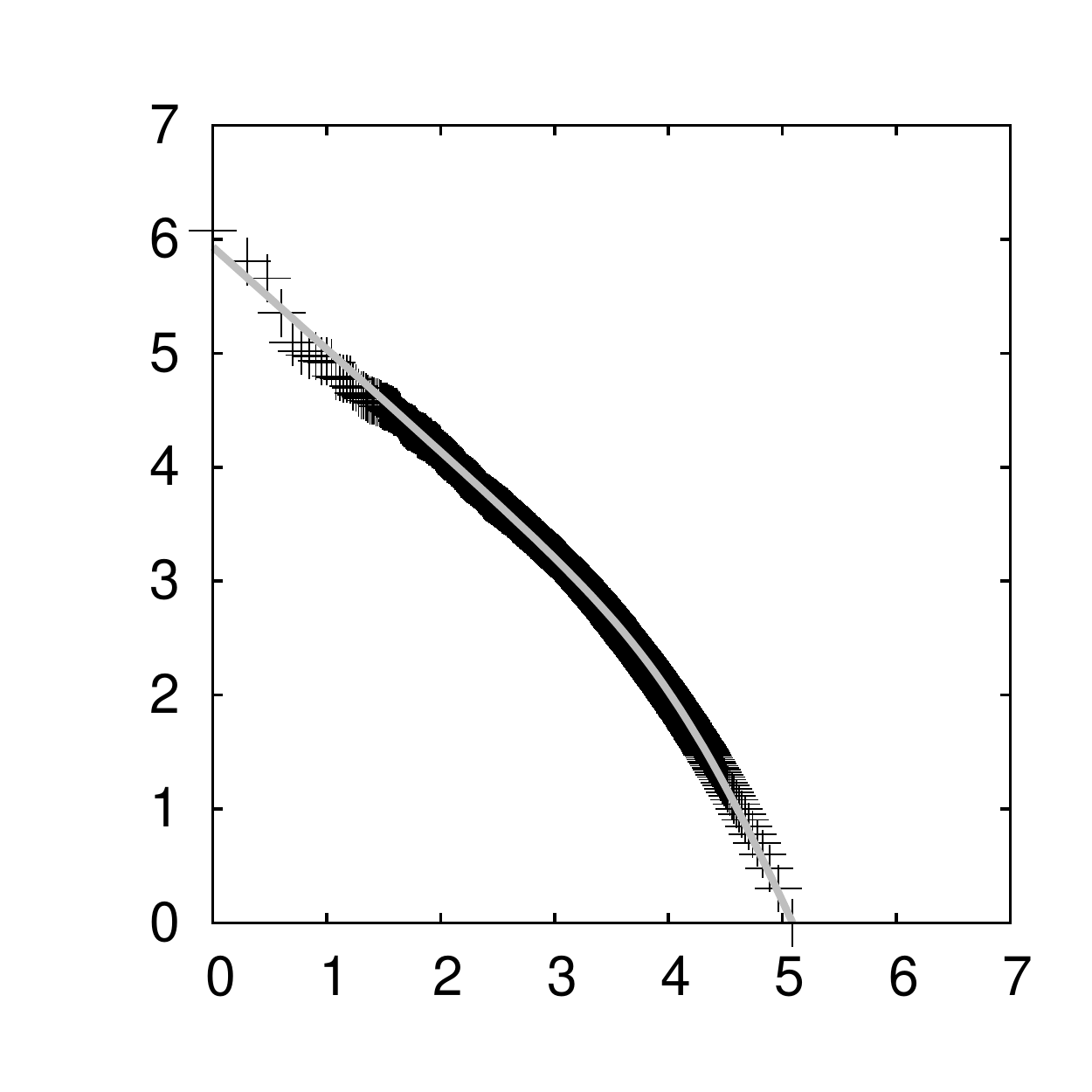}
\end{minipage}
\vspace{-3mm}
\caption{four-parameter (left) and two-parameter (right) fitting on {\tt lt}.}
\vspace{-3mm}
\label{fig:lt}
\end{figure}

\begin{figure}[t]
\centering
\begin{minipage}{0.45\linewidth}
    \centering
    \includegraphics[width=1.0\linewidth]{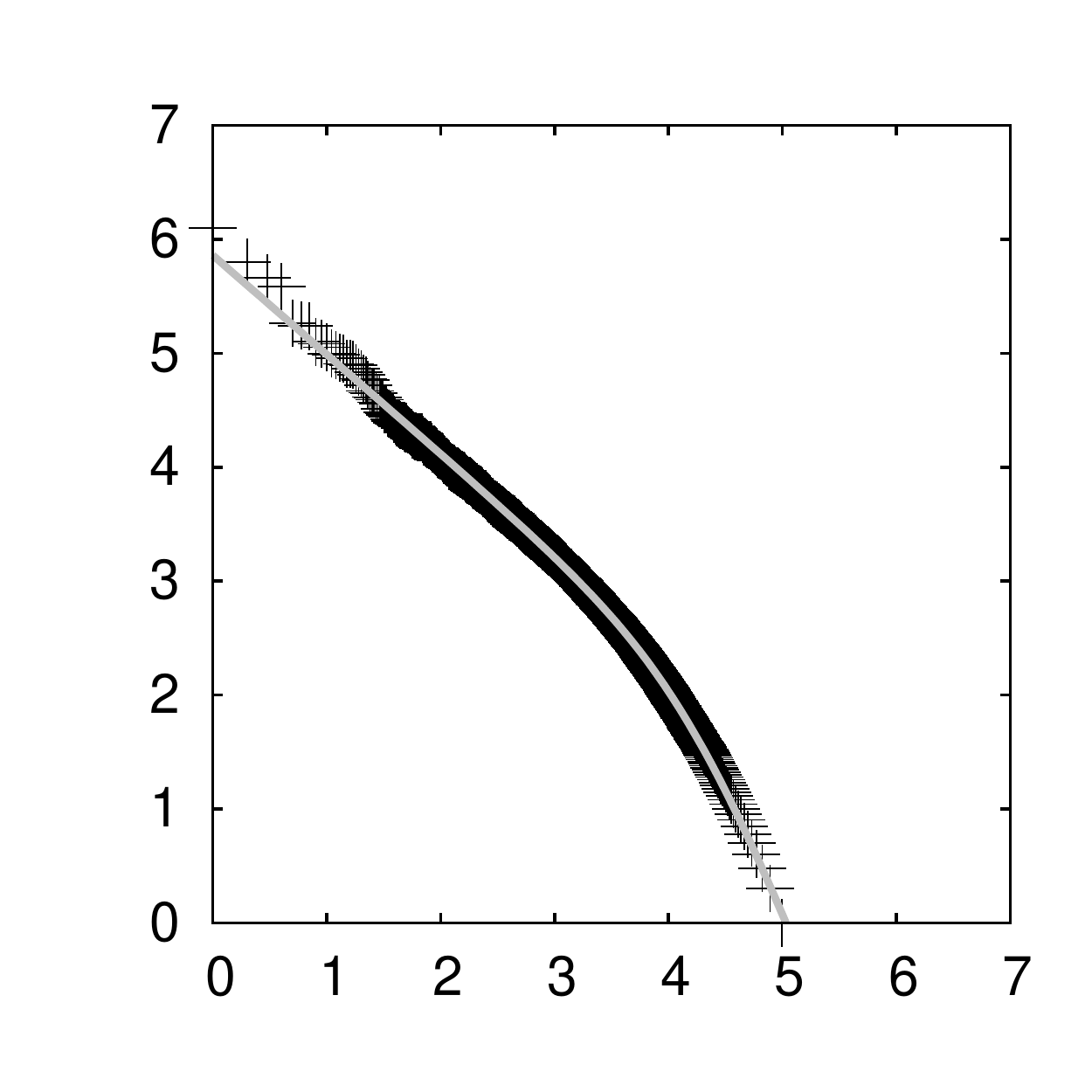}
\end{minipage}
\begin{minipage}{0.45\linewidth}
    \centering
    \includegraphics[width=1.0\linewidth]{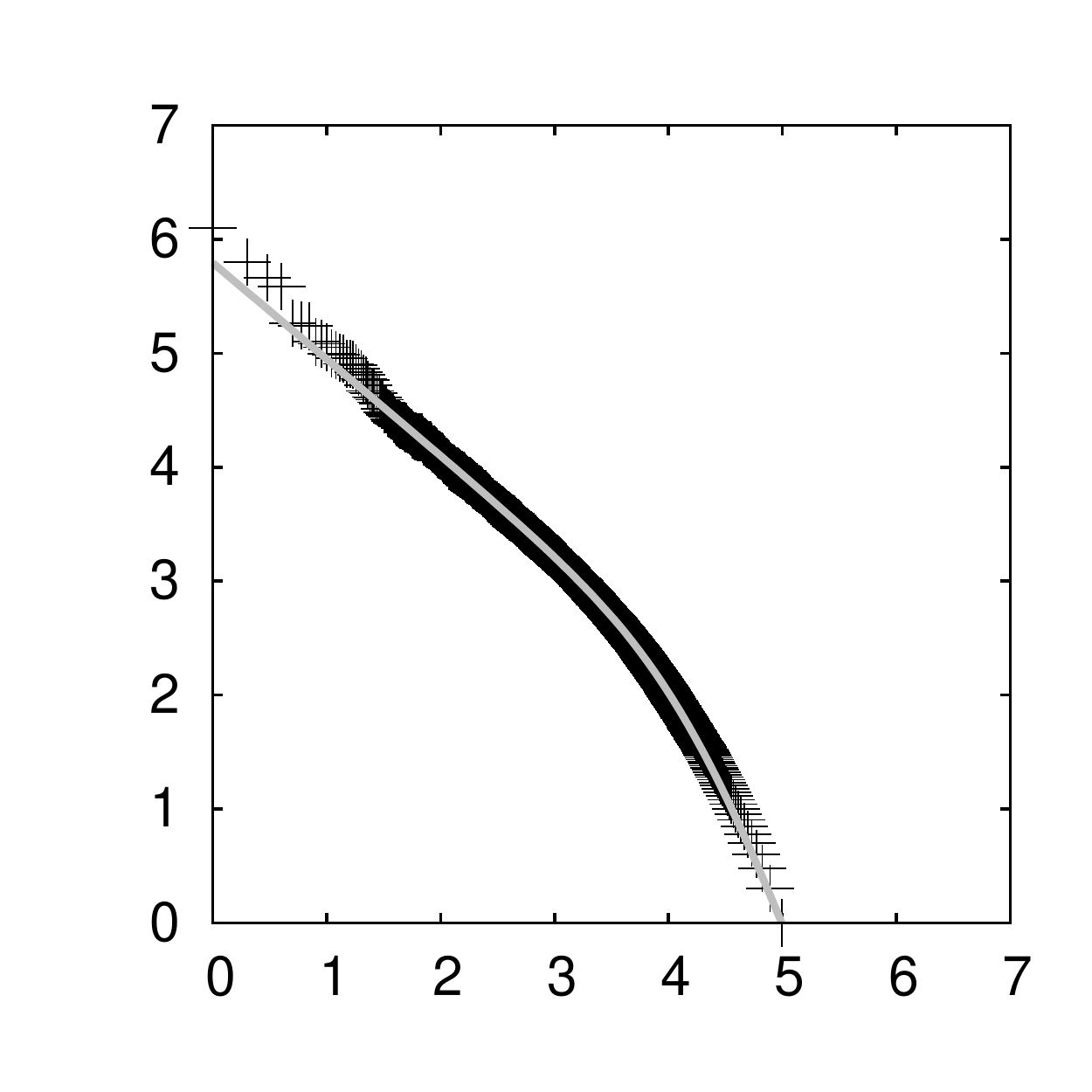}
\end{minipage}
\vspace{-3mm}
\caption{four-parameter (left) and two-parameter (right) fitting on {\tt lv}.}
\vspace{-3mm}
\label{fig:lv}
\end{figure}

\begin{figure}[t]
\centering
\begin{minipage}{0.45\linewidth}
    \centering
    \includegraphics[width=1.0\linewidth]{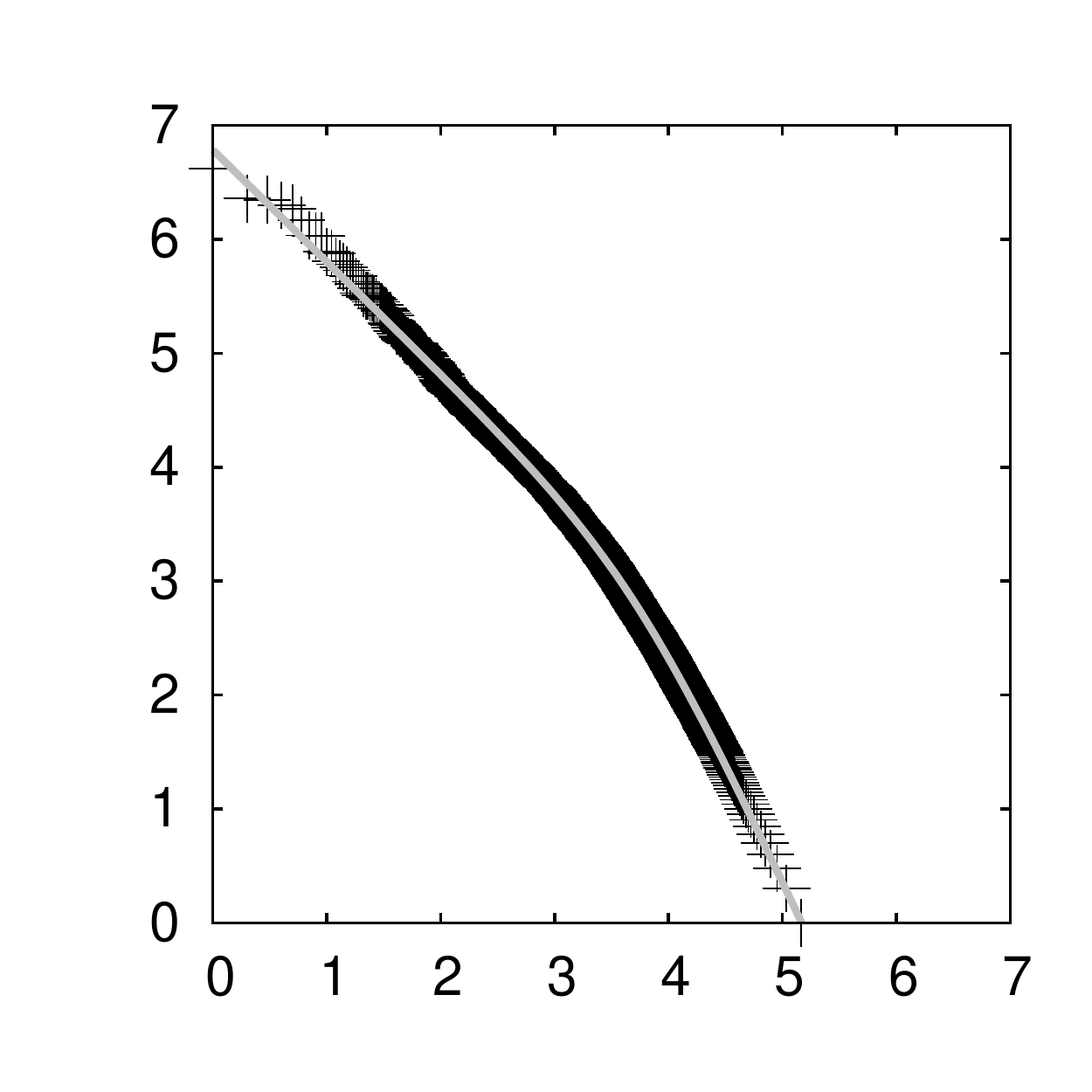}
\end{minipage}
\begin{minipage}{0.45\linewidth}
    \centering
    \includegraphics[width=1.0\linewidth]{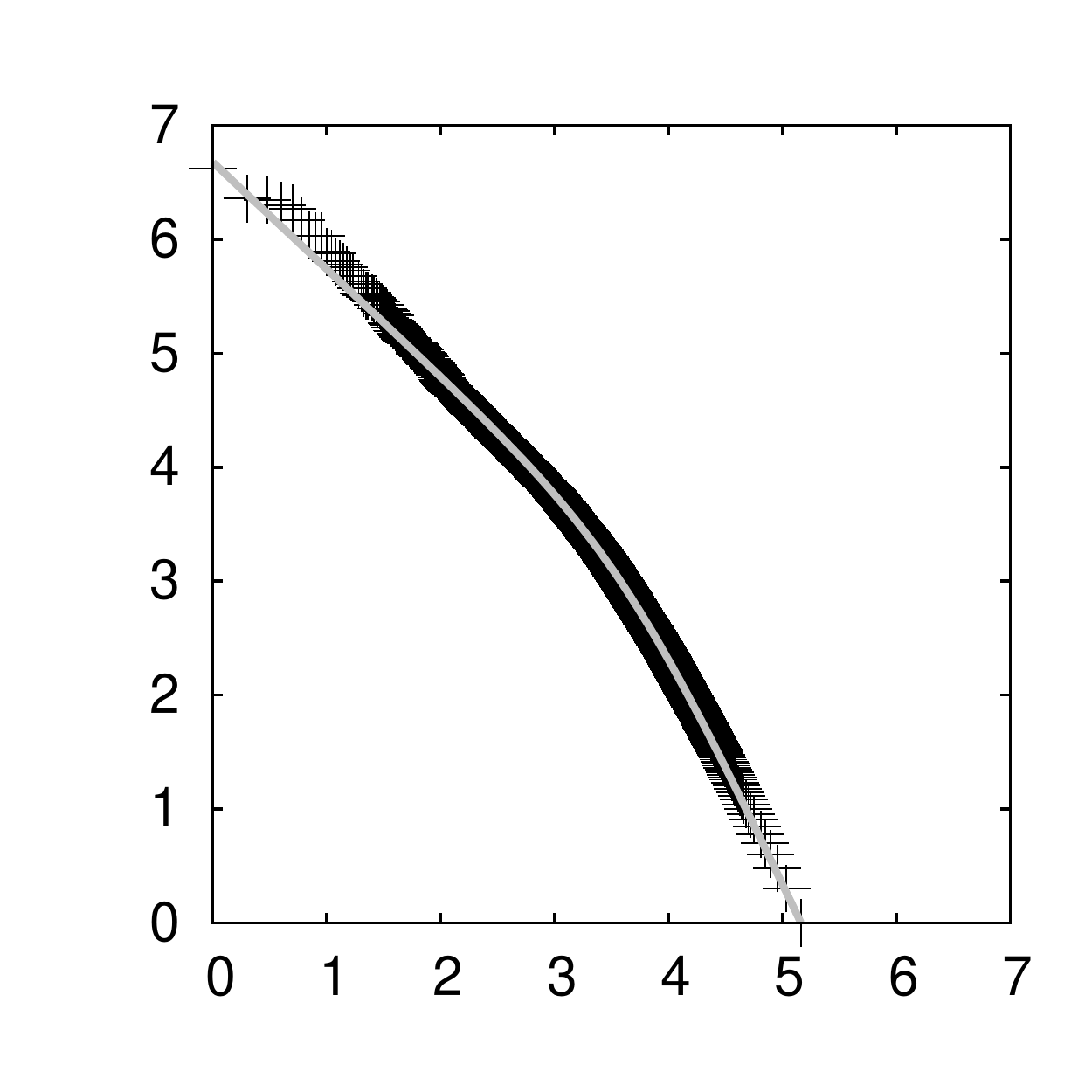}
\end{minipage}
\vspace{-3mm}
\caption{four-parameter (left) and two-parameter (right) fitting on {\tt nl}.}
\vspace{-3mm}
\label{fig:nl}
\end{figure}

\begin{figure}[t]
\centering
\begin{minipage}{0.45\linewidth}
    \centering
    \includegraphics[width=1.0\linewidth]{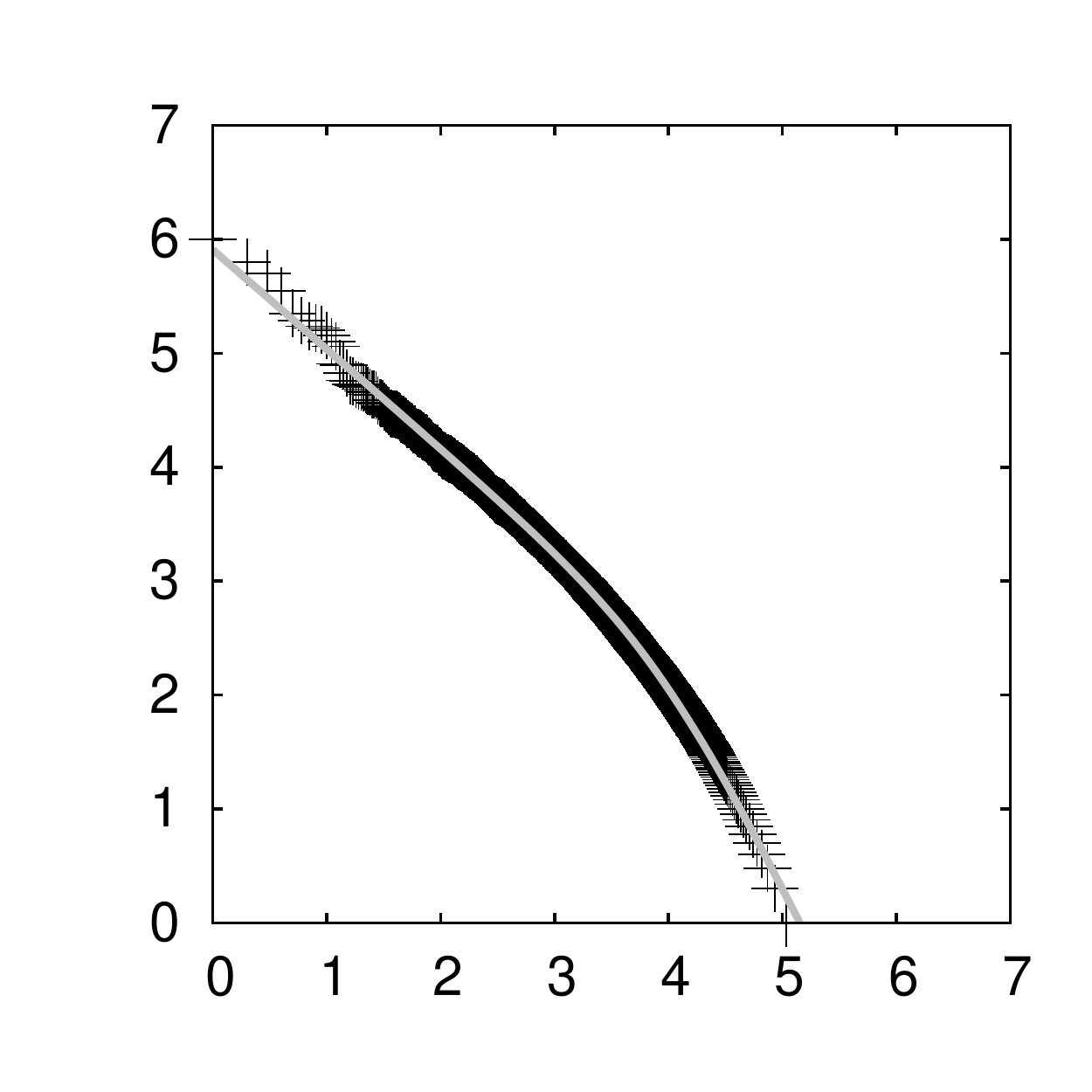}
\end{minipage}
\begin{minipage}{0.45\linewidth}
    \centering
    \includegraphics[width=1.0\linewidth]{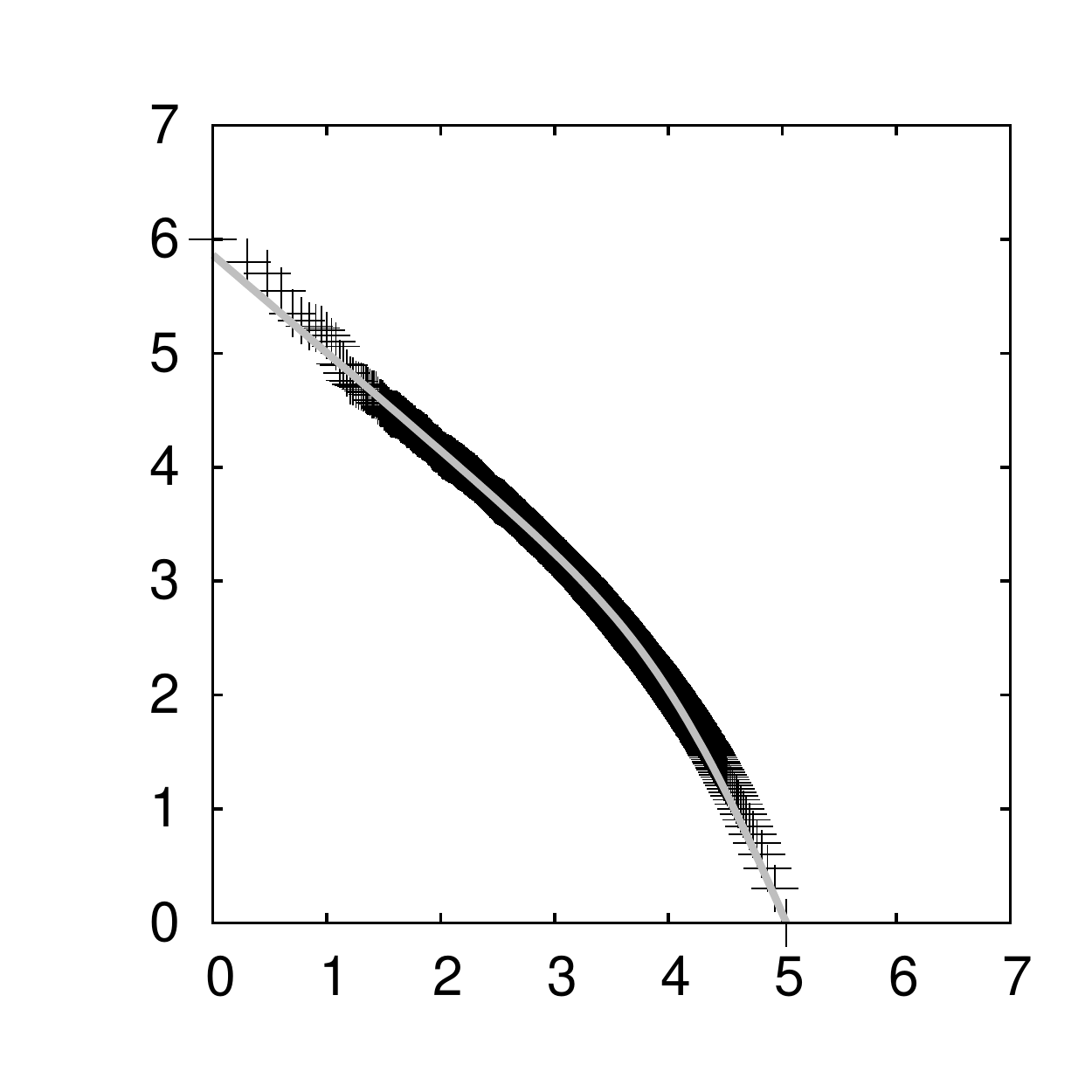}
\end{minipage}
\vspace{-3mm}
\caption{four-parameter (left) and two-parameter (right) fitting on {\tt pl}.}
\vspace{-3mm}
\label{fig:pl}
\end{figure}

\begin{figure}[t]
\centering
\begin{minipage}{0.45\linewidth}
    \centering
    \includegraphics[width=1.0\linewidth]{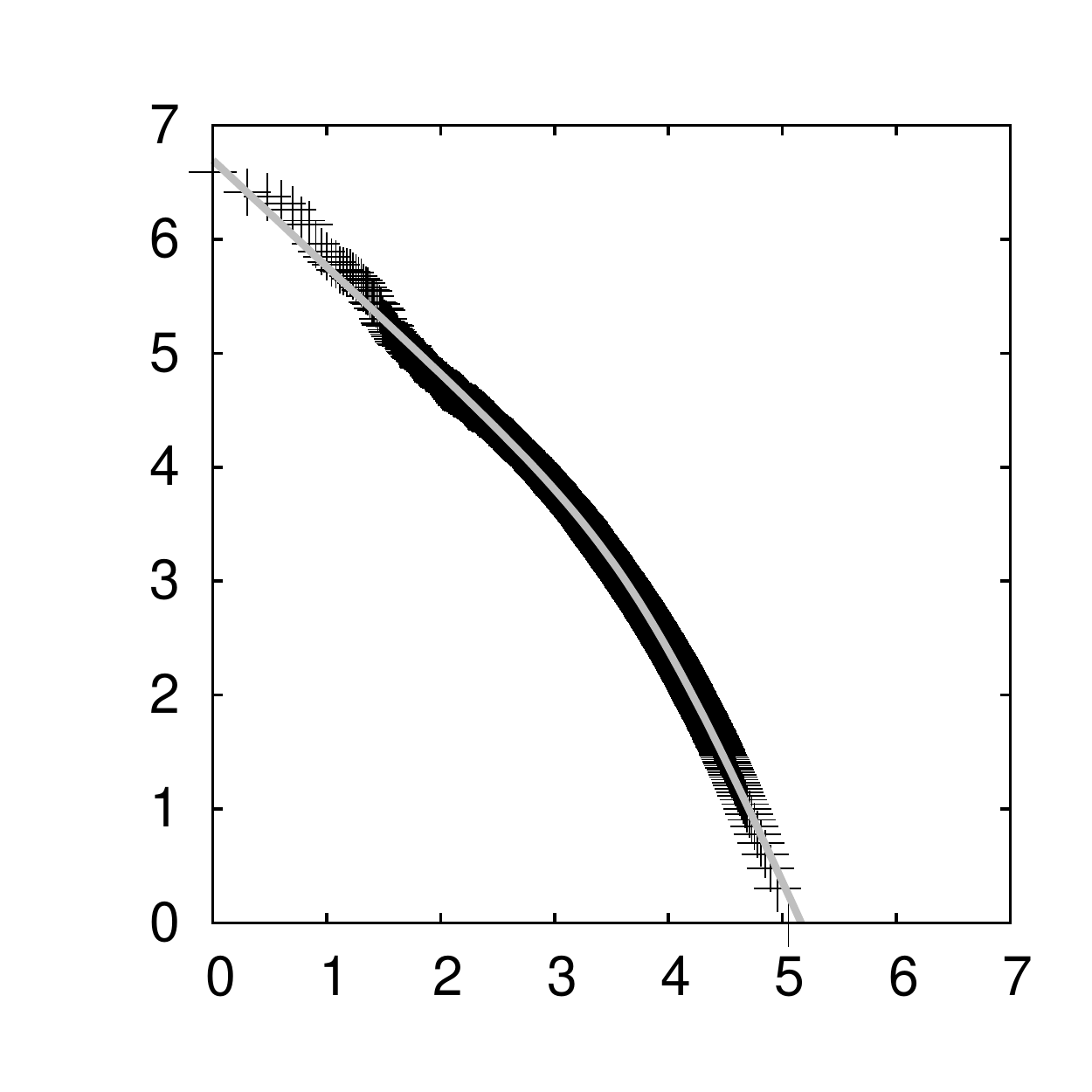}
\end{minipage}
\begin{minipage}{0.45\linewidth}
    \centering
    \includegraphics[width=1.0\linewidth]{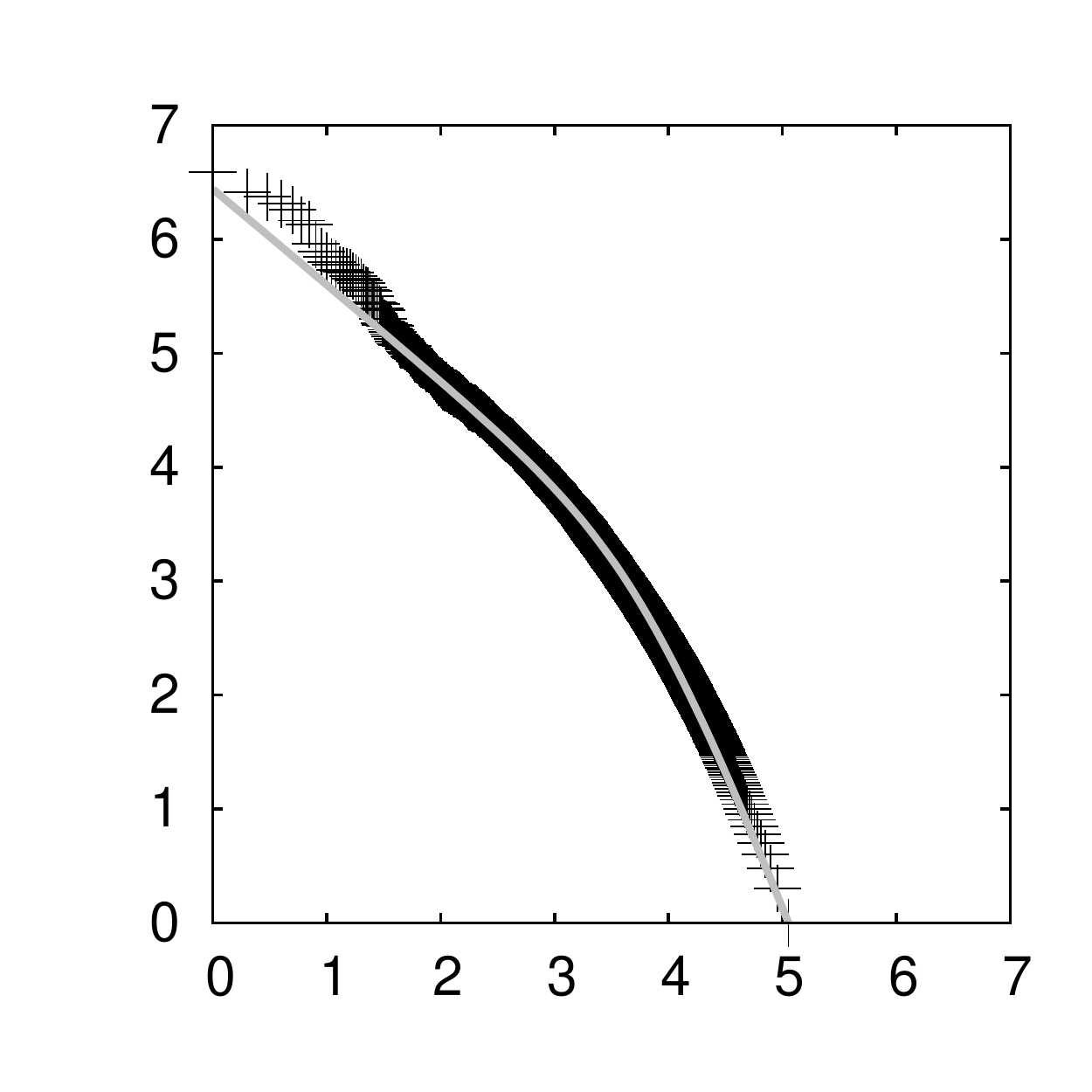}
\end{minipage}
\vspace{-3mm}
\caption{four-parameter (left) and two-parameter (right) fitting on {\tt pt}.}
\vspace{-3mm}
\label{fig:pt}
\end{figure}

\begin{figure}[t]
\centering
\begin{minipage}{0.45\linewidth}
    \centering
    \includegraphics[width=1.0\linewidth]{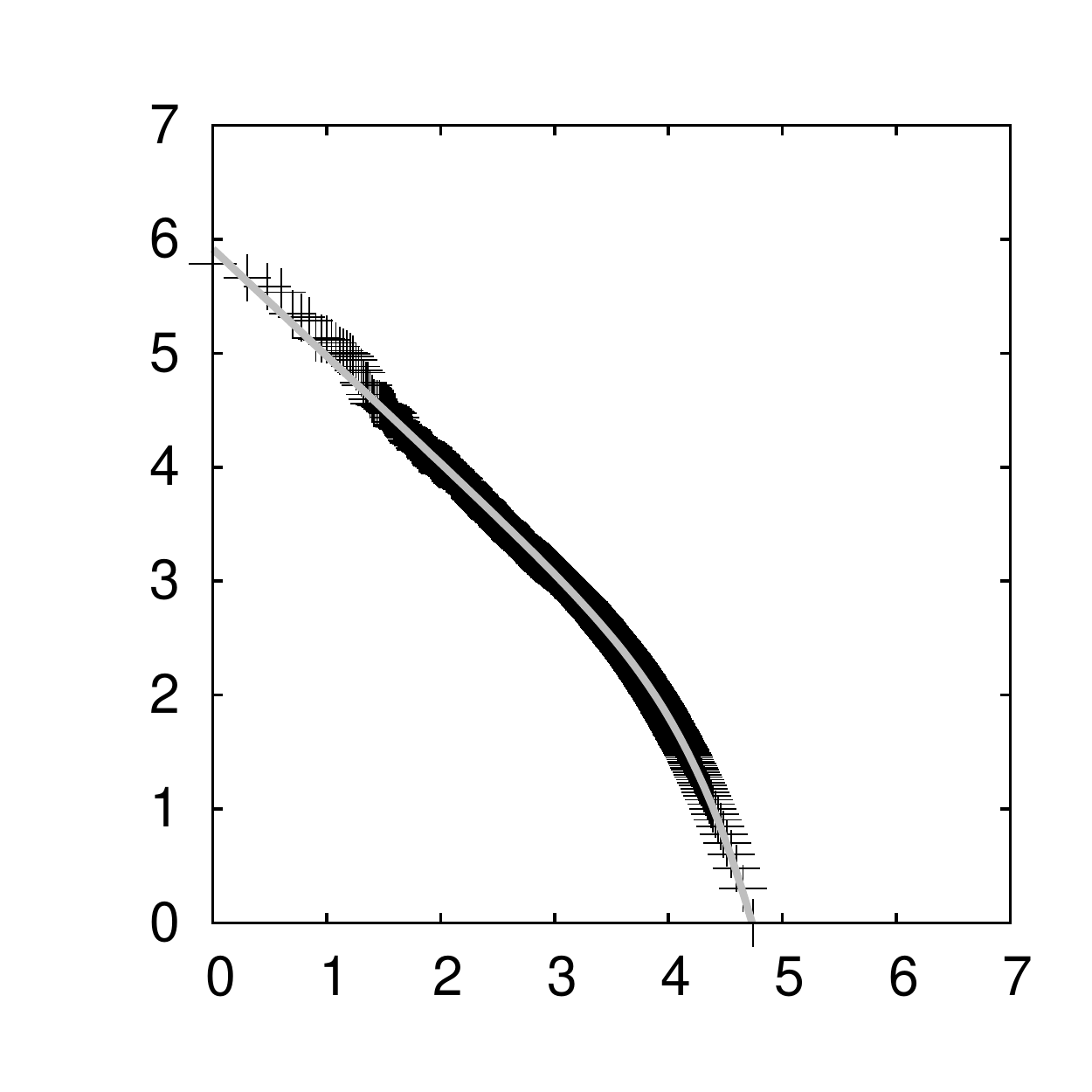}
\end{minipage}
\begin{minipage}{0.45\linewidth}
    \centering
    \includegraphics[width=1.0\linewidth]{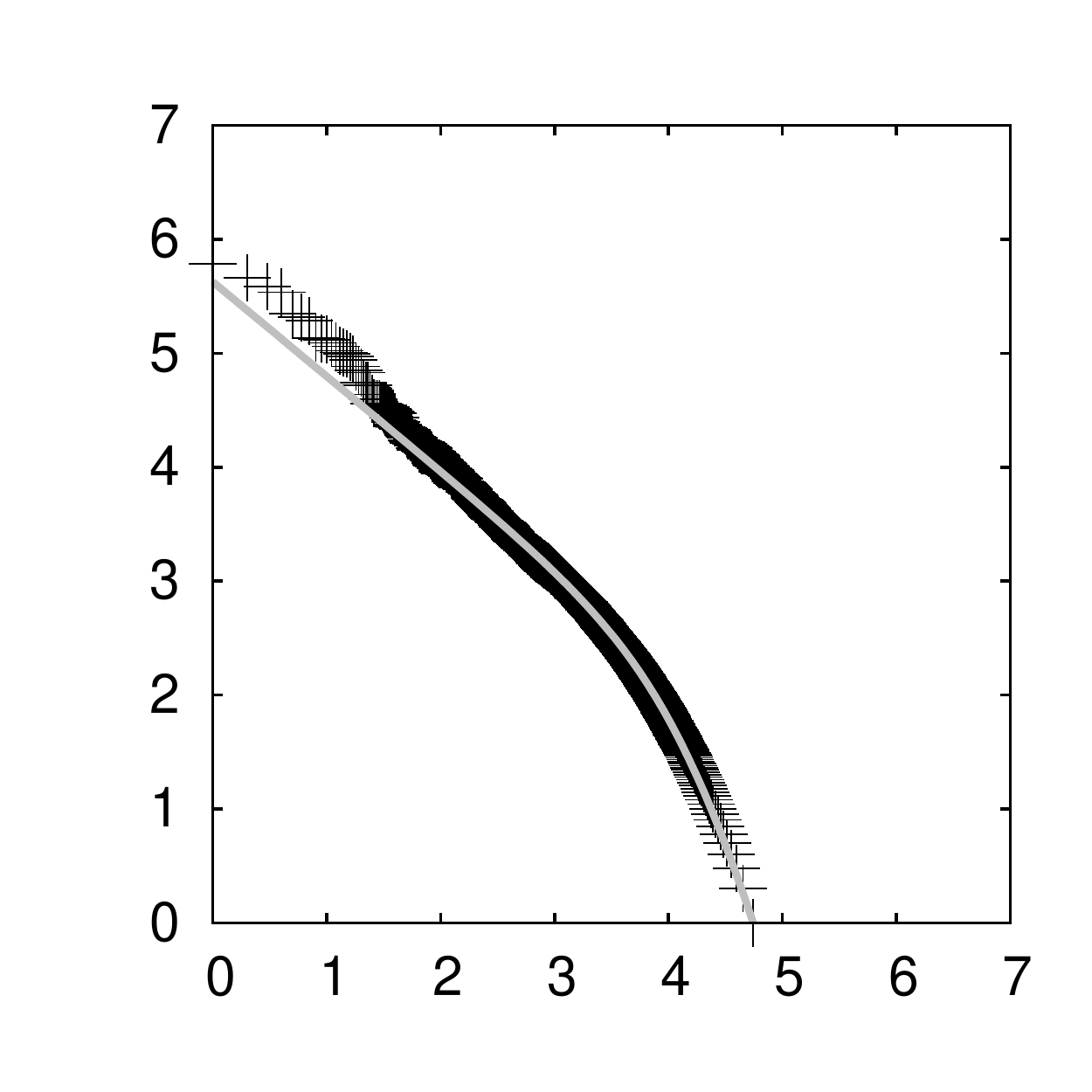}
\end{minipage}
\vspace{-3mm}
\caption{four-parameter (left) and two-parameter (right) fitting on {\tt ro}.}
\vspace{-3mm}
\label{fig:ro}
\end{figure}

\begin{figure}[t]
\centering
\begin{minipage}{0.45\linewidth}
    \centering
    \includegraphics[width=1.0\linewidth]{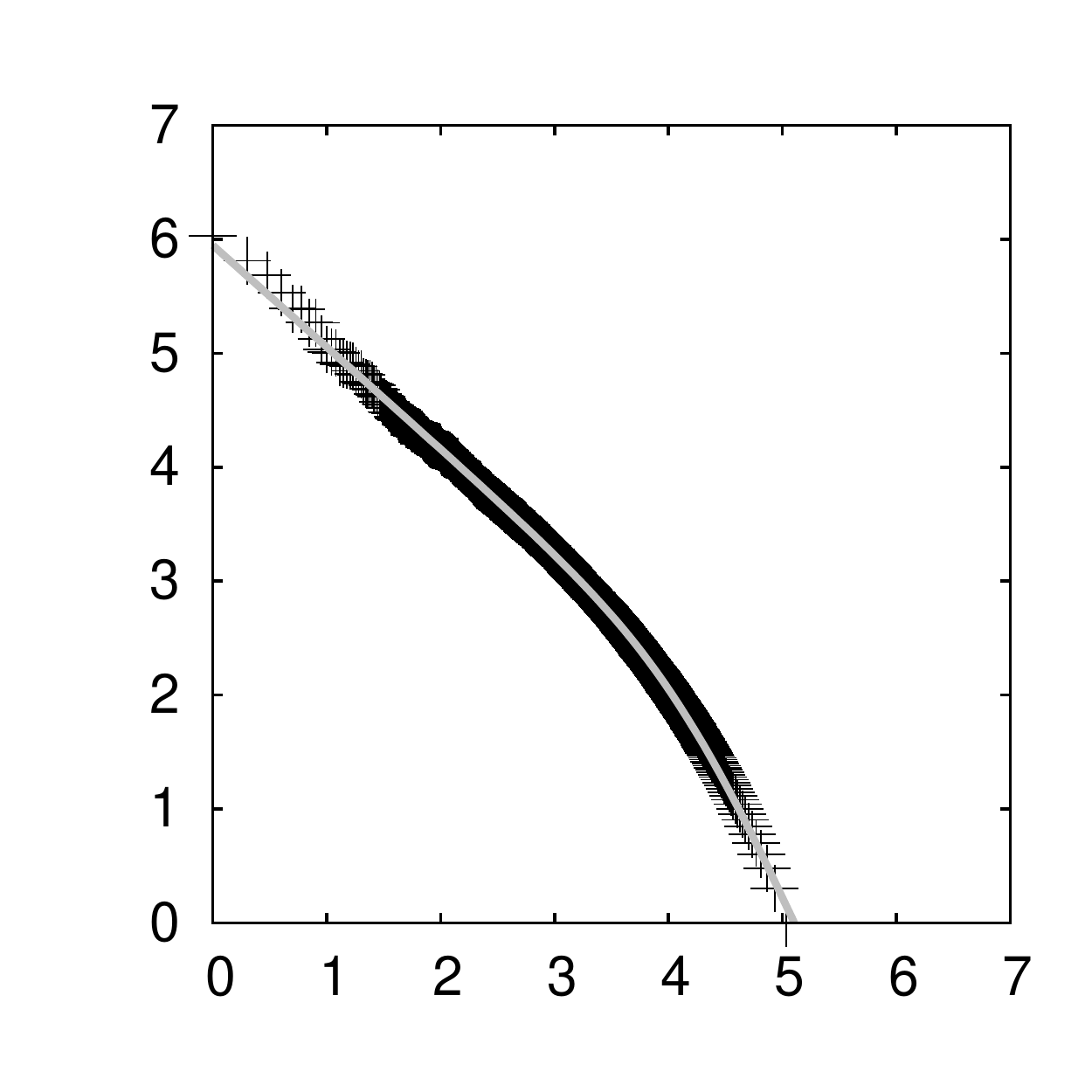}
\end{minipage}
\begin{minipage}{0.45\linewidth}
    \centering
    \includegraphics[width=1.0\linewidth]{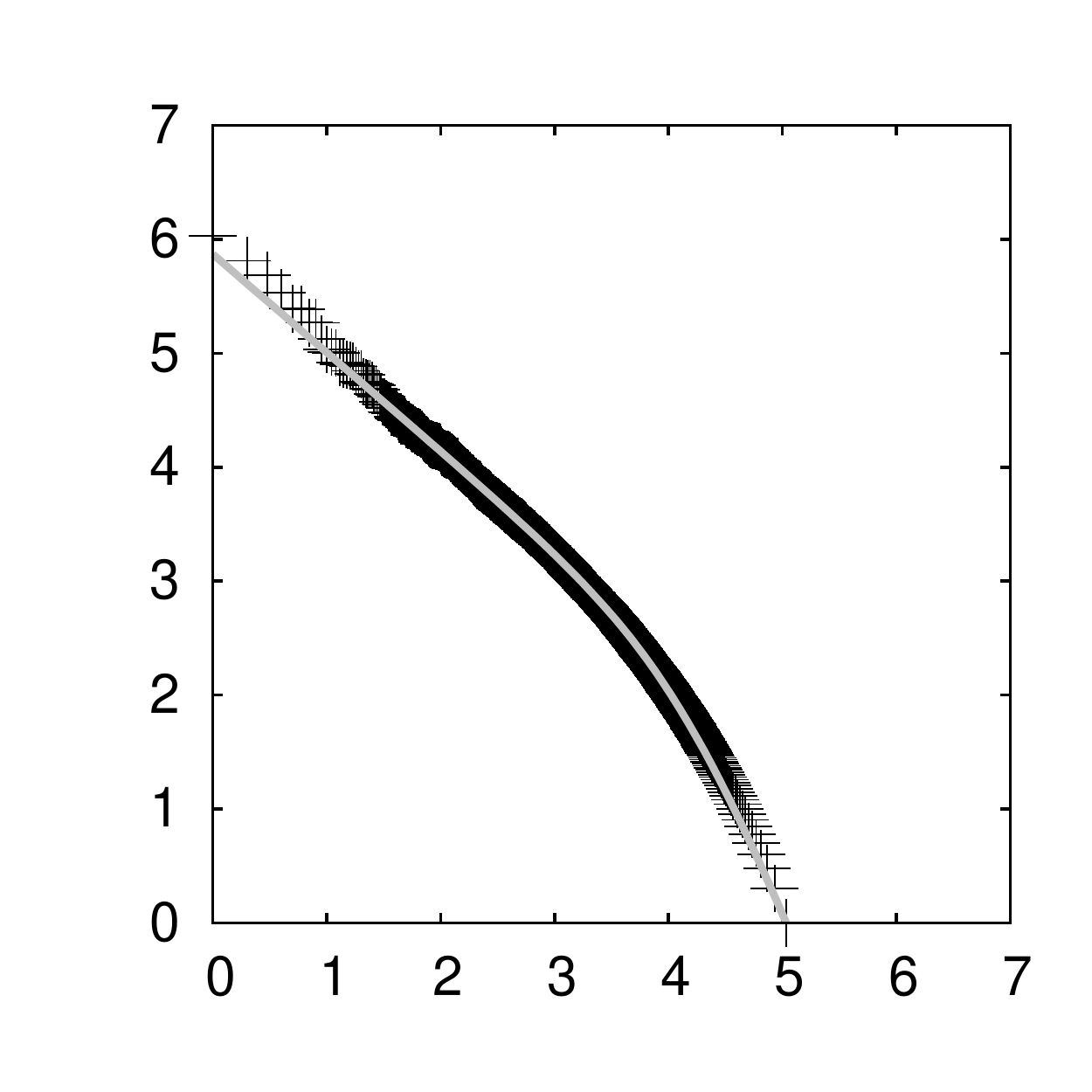}
\end{minipage}
\vspace{-3mm}
\caption{four-parameter (left) and two-parameter (right) fitting on {\tt sk}.}
\vspace{-3mm}
\label{fig:sk}
\end{figure}

\begin{figure}[t]
\centering
\begin{minipage}{0.45\linewidth}
    \centering
    \includegraphics[width=1.0\linewidth]{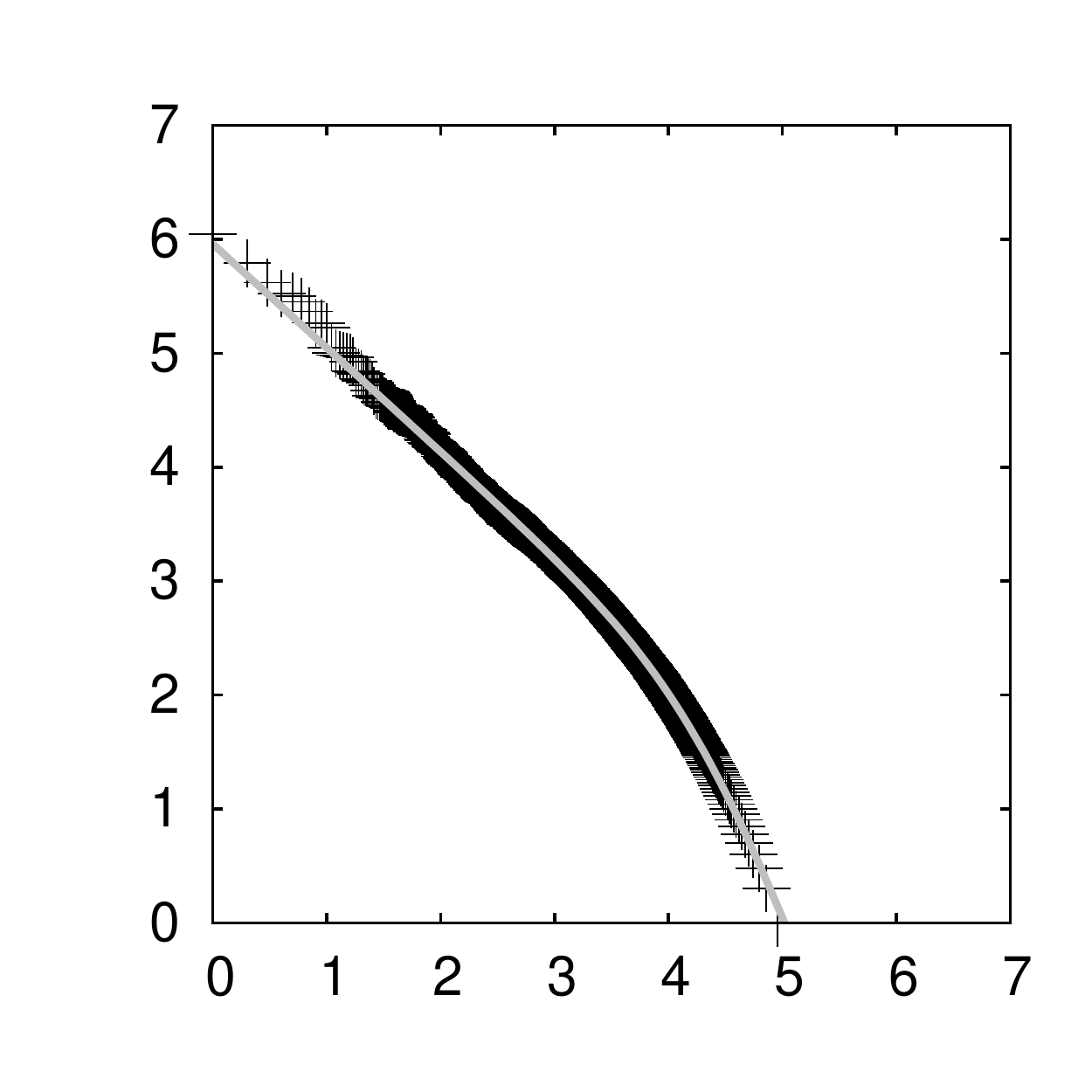}
\end{minipage}
\begin{minipage}{0.45\linewidth}
    \centering
    \includegraphics[width=1.0\linewidth]{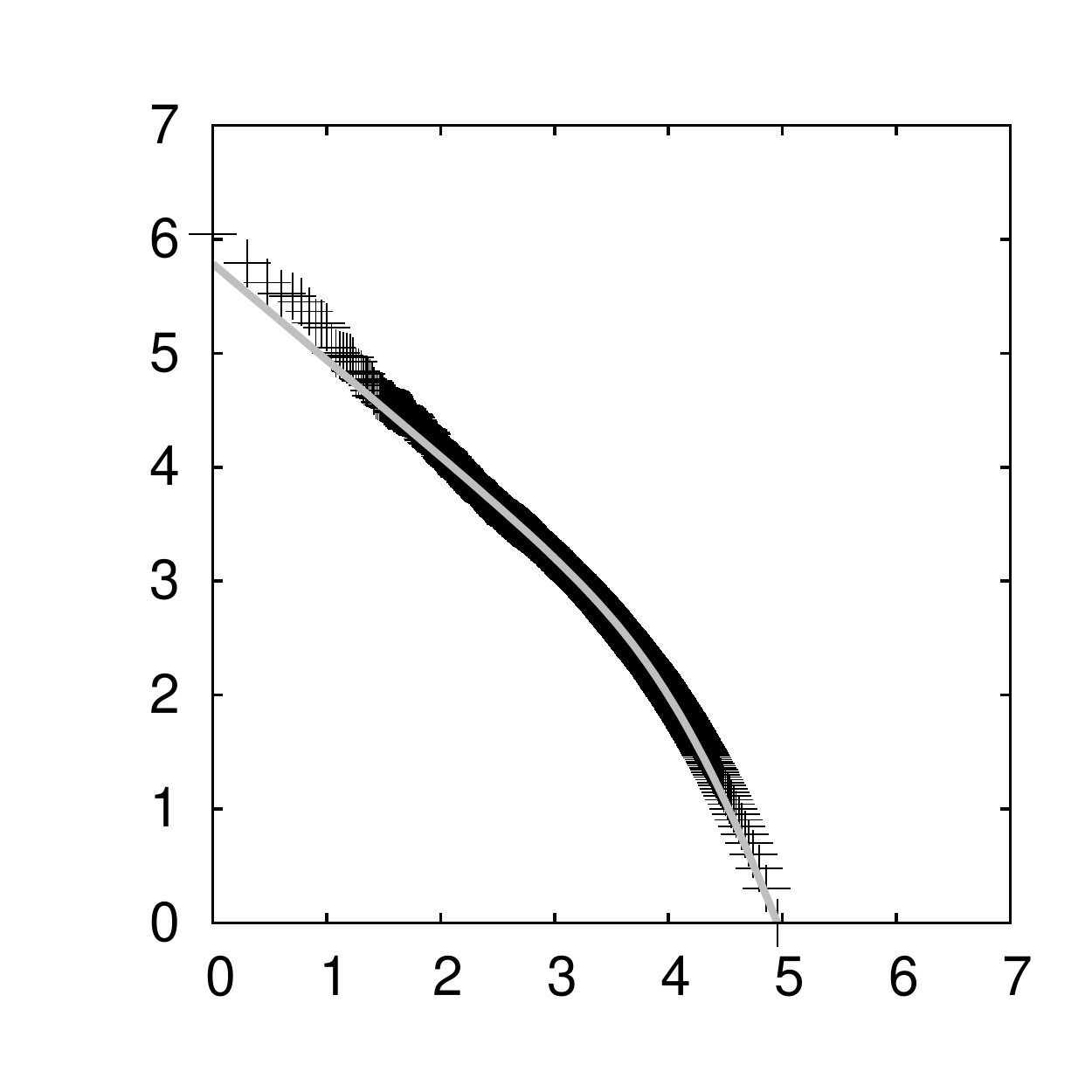}
\end{minipage}
\vspace{-3mm}
\caption{four-parameter (left) and two-parameter (right) fitting on {\tt sl}.}
\vspace{-3mm}
\label{fig:sl}
\end{figure}

\begin{figure}[t]
\centering
\begin{minipage}{0.45\linewidth}
    \centering
    \includegraphics[width=1.0\linewidth]{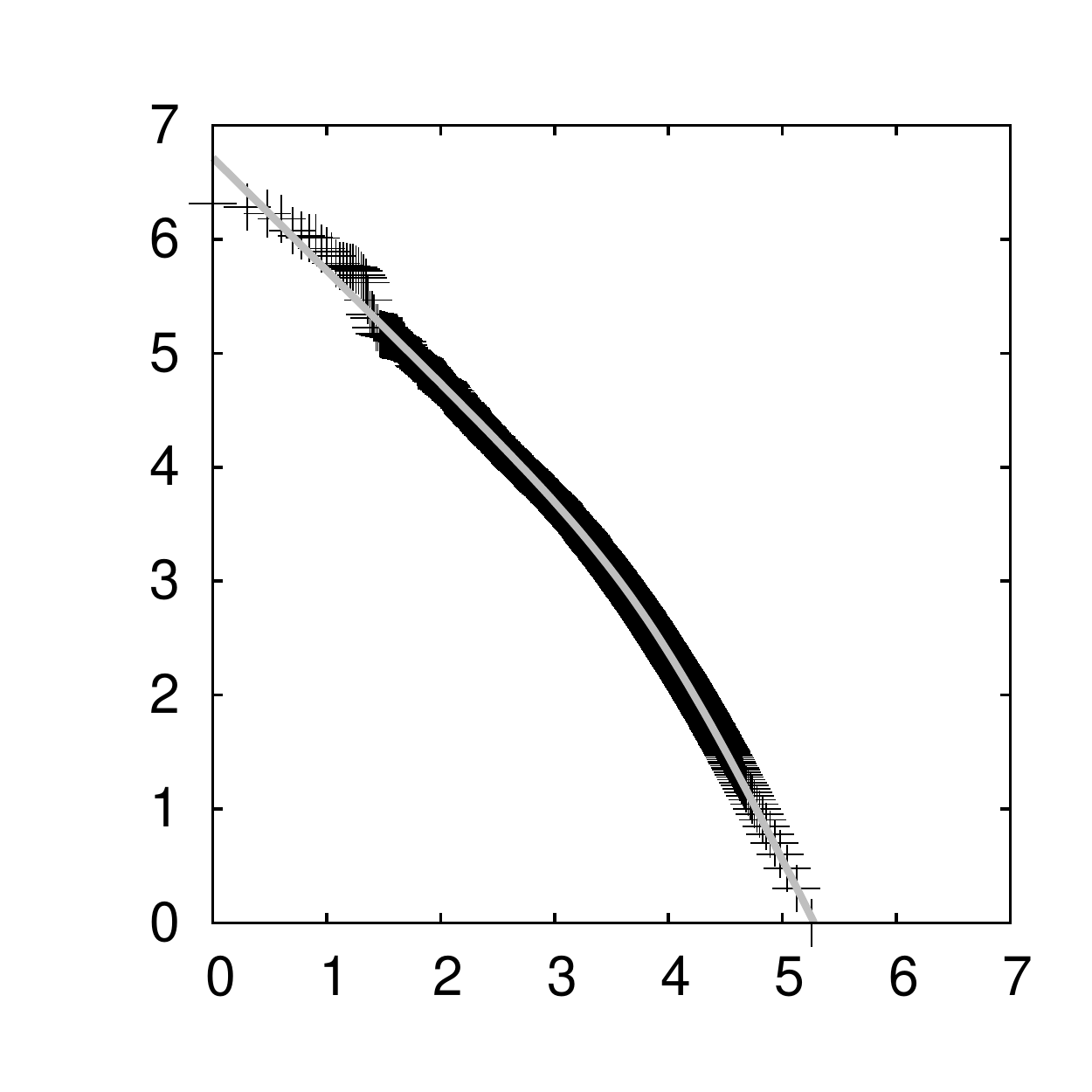}
\end{minipage}
\begin{minipage}{0.45\linewidth}
    \centering
    \includegraphics[width=1.0\linewidth]{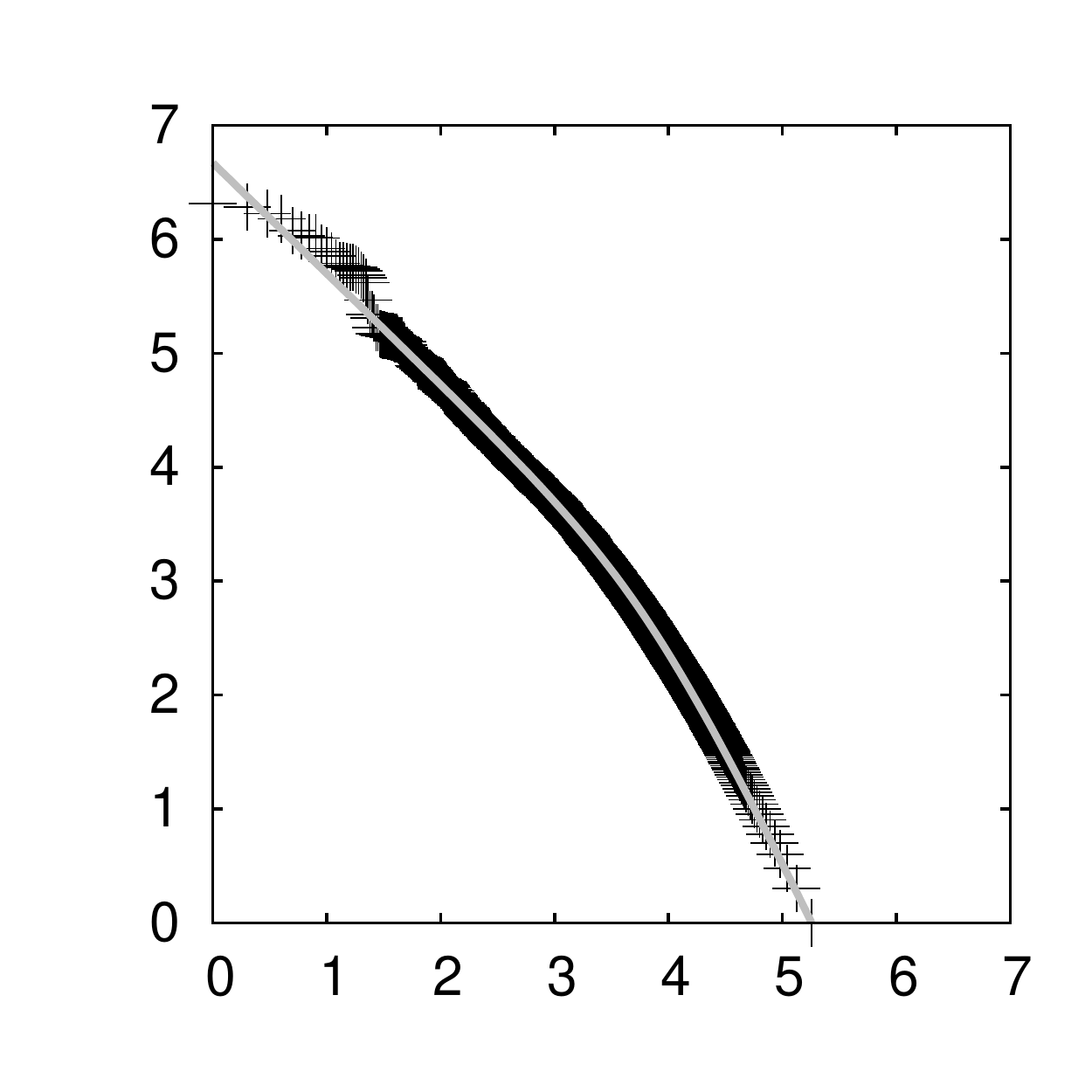}
\end{minipage}
\vspace{-3mm}
\caption{four-parameter (left) and two-parameter (right) fitting on {\tt sv}.}
\vspace{-3mm}
\label{fig:sv}
\end{figure}

\end{document}